\documentclass[journal]{IEEEtran}

\usepackage{array}
\usepackage[caption=false,font=normalsize,labelfont=sf,textfont=sf]{subfig}
\usepackage{caption}
\usepackage{textcomp}
\usepackage{stfloats}
\usepackage{url}
\usepackage{mathtools}
\usepackage{verbatim}
\def\BibTeX{{\rm B\kern-.05em{\sc i\kern-.025em b}\kern-.08em
    T\kern-.1667em\lower.7ex\hbox{E}\kern-.125emX}}
\usepackage{balance}
\usepackage{multirow}
\usepackage{booktabs}
\usepackage{algorithmic}%
\usepackage{algorithm}
\usepackage{rotating}
\usepackage{color}
\usepackage{graphicx}
\usepackage{subcaption}
\usepackage{amsmath,amssymb,amsfonts}
\usepackage{pifont}
\newcolumntype{C}[1]{>{\centering\arraybackslash}m{#1}}
\newcommand{\cmark}{\ding{51}} % Checkmark
\newcommand{\xmark}{\ding{55}} % Cross
\usepackage[colorlinks,urlcolor=blue,linkcolor=blue,citecolor=blue]{hyperref}
\usepackage{cite}
\usepackage{lettrine}

\makeatletter
\renewcommand{\subsection}{\@startsection{subsection}{2}{\z@}%
  {1.0ex plus 0.4ex minus 0.2ex}
  {0.6ex plus 0.3ex}
  {\normalfont\normalsize\itshape}}%
\makeatother

\begin{document}

\title{GeNeX: Genetic Network eXperts framework for
addressing Validation Overfitting}

\author{Emmanuel Pintelas and Ioannis E. Livieris.
% % %
\thanks{E. Pintelas, is with the Department of Statistics \& Insurance Science, University of Piraeus, GR 18534, Greece (e-mail: e.pintelas@unipi.gr).}
\thanks{I.E. Livieris, is with the Department of Business \& Organization Administration, University of Peloponnese, GR 24100, Greece (e-mail: livieris@uop.gr).}
% % 
}

\markboth{Accepted for publication in IEEE TRANSACTIONS ON NEURAL NETWORKS AND LEARNING SYSTEMS}%
{How to Use the IEEEtran \LaTeX \ Templates}

\maketitle

{\small\textit{This work has been accepted for publication in IEEE 
		Transactions on Neural Networks and Learning Systems. \copyright~IEEE. 
		Personal use of this material is permitted. Permission from IEEE must 
		be obtained for all other uses, in any current or future media, 
		including reprinting/republishing this material for advertising or 
		promotional purposes, creating new collective works, for resale or 
		redistribution to servers or lists, or reuse of any copyrighted 
		component of this work in other works.}}
	
\begin{abstract}

Excessive reliance on validation performance during model selection can lead to validation overfitting (VO), where models appear effective during development but fail at test time. This issue is further amplified in low-data regimes and under distribution shifts, where validation signals become unreliable. Although ensemble learning is widely used to improve robustness and generalization, most ensemble construction pipelines depend heavily on validation scores, leaving them vulnerable to VO and limiting their reliability in real-world deployment scenarios.
To address this, we propose GeNeX (Genetic Network eXperts), a framework that mitigates validation overfitting at both model generation and ensemble construction stages. 
In the generation phase, GeNeX employs a dual-path strategy: gradient-based training is coupled with genetic model evolution. Offspring networks are created via crossover of trained parents, promoting structural diversity and weight-level regeneration without relying on validation feedback. This results in a candidate pool of robust, non-overfitted models. 
During ensemble construction, the candidate networks are clustered by prediction behavior to identify complementary model spaces. Within each cluster, multiple diverse experts are selected using criteria such as robustness and representativeness, and fused at the weight level to form compact prototype networks. The final ensemble aggregates these prototypes, with predictions optimized via Sequential Quadratic Programming for output-level synergy. 
To rigorously evaluate VO resilience, we introduce a VO-aware evaluation protocol that simulates realistic deployment scenarios by enforcing distributional divergence between training and test subsets. Experiments on four real-world image classification tasks (Skin Cancer, Deepfake Detection, Plant Disease, Pneumonia) demonstrate that GeNeX consistently outperforms state-of-the-art ensembles under limited-data and shift-aware conditions, achieving strong generalization with minimal VO gaps.

\end{abstract}


\section{Introduction}

\IEEEPARstart{V}{alidation} overfitting (VO) consists of an underexplored, yet crucial challenge in deep learning model development \cite{pintelas2025mobilenet,zhao2024breaking}. Unlike common training overfitting, which occurs when a model memorizes the training data and fails to generalize, VO arises when a model appears optimal based on validation metrics but ultimately performs poorly on truly unseen test data—particularly under distribution shifts where validation and test distributions diverge \cite{zhao2024breaking,xu2024cbdmoe}. This phenomenon is typically caused by excessive reliance on a limited or non-representative validation set during model selection, hyperparameter tuning, or ensemble construction \cite{sarmah2024ensemble}, leading to inflated validation scores that poorly reflect real-world performance.

The risk of validation overfitting increases significantly in scenarios characterized by repeated evaluations on the same validation set, particularly under limited-data conditions or in the presence of distribution shifts \cite{zhao2024breaking,xu2024cbdmoe}. In such cases, models may exploit statistical flukes in the validation set, leading to misleadingly high validation scores without true generalization capabilities \cite{pintelas2025mobilenet,pintelas2025textnex}. While validation monitoring is traditionally employed to avoid training overfitting and select best-validation checkpoint, heavy dependence on validation feedback acts as a double-edged sword: it mitigates training overfitting but simultaneously increases the risk of validation overfitting. Consequently, even models exhibiting small train-validation gaps and high validation accuracy may collapse when evaluated on new data \cite{zhao2024breaking}.

This issue is particularly prevalent in real-world deployments with small or biased validation sets, research benchmarks in low-resource scenarios and competition-driven environments \cite{xu2024cbdmoe}. Despite its significance, validation overfitting remains rarely addressed directly in deep learning pipelines, 
where model selection often continues to favor validation-optimized candidates. Such practices lead to the development of models that, despite strong validation performance, fail catastrophically when deployed \cite{zhao2024breaking}.

Recent research has explored ensemble strategies aiming at improving generalization and robustness, including methods, which promote expert diversity \cite{zhao2024breaking,xu2024cbdmoe}, employ genetic optimization search \cite{xu2024genetic,kaya2024novel,dar2024deep}, or cluster models to identify experts based on predictive behavior \cite{pintelas2025mobilenet,pintelas2025textnex}. While these methods enhance diversity or structural complementarity, they often rely heavily on validation feedback during model selection. Moreover, most fail to explicitly control VO risk during model generation or ensemble construction. As a result, they remain vulnerable to VO, particularly in low-resource settings or under significant test-time distribution shifts—conditions commonly encountered in real-world deployments \cite{zhao2024breaking}.

To address these limitations, we propose GeNeX (Genetic Network eXperts), an ensemble framework designed explicitly to minimize the risk of validation overfitting while maintaining high generalization and robustness. GeNeX consists of two core modules: (i) a Genetic-based Ensemble generator \textit{(GenE)}, which generates a diverse population of robust networks by combining supervised training and genetic weight fusion, with zero reliance on validation feedback; and (ii) Prototype-based eXperts selector \textit{(ProtoNeX)}, which clusters the generated networks based on behavioral similarity, elects diverse experts within each cluster, and synthesizes prototype models through weight-level aggregation.

Our hypothesis is that incorporating a genetic-based weight-crossover mechanism into a standard gradient-based optimization pipeline can act as a ``refreshing'' regeneration process, mitigating overfitting in the networks' weight level, while promoting diversity and robustness without validation dependency; thereby, drastically reducing VO risk. Furthermore, constructing the ensemble via prototype-based expert fusion ensures complementary representation across the model space, enhancing the final ensemble's stability and performance.

In order to rigorously validate this hypothesis, we designed a VO-aware evaluation protocol, which simulates realistic validation overfitting risks by deliberately enforcing distributional shift between training and testing sets. Rather than relying on random sampling, we introduce a Jensen-Shannon Divergence (JSD)-guided partitioning algorithm that maximizes divergence between subsets while maintaining a predefined user data ratio (e.g., 30–70), enabling researchers to create VO-prone scenarios in any case-study dataset. JSD provides a symmetric and bounded measure of divergence, making it a robust choice for quantifying distributional differences \cite{englesson2021generalized}. In our protocol, JSD specifically acts as the optimization criterion during dataset splitting, and subsequently serves as a direct indicator of VO susceptibility per task. High JSD implies strong distributional shifts between training/validation and testing distributions, indicating greater VO risk. Therefore, models that achieve high predictive performance while maintaining a low VO gap under high-JSD conditions demonstrate enhanced robustness and credibility in addressing VO.

Employing this controlled experimental setup, evaluation across four diverse, real-world image classification case studies consistently demonstrates GeNeX's superior ability to achieve high predictive performance while effectively minimizing validation overfitting. Under challenging high-JSD scenarios, GeNeX significantly outperforms state-of-the-art ensemble baselines, underscoring its robustness and generalizability in mitigating VO. 

The main contributions of this research work are summarized as follows:
\begin{itemize}
    \item We formally analyze the phenomenon of validation overfitting, quantifying its causes from both modeling and data perspectives.
    \item We propose GeNeX, a two-phase ensemble learning framework that minimizes validation overfitting by combining genetic-based model generation with prototype-based clustering and fusion, achieving robust and high generalization performance under high VO risk.
    \item We introduce a VO-aware evaluation protocol built on a dedicated JSD-guided data partitioning algorithm to simulate realistic distributional shifts for benchmarking VO resilience, offering significant utility for both academic research and industrial deployment pipelines. 
\end{itemize}

\section{Definition of Validation Overfitting \& Motivation}\label{vo_definition}

Validation overfitting (VO) refers to the phenomenon where a model exhibits strong performance on a validation set during development but fails to generalize to truly unseen test data \cite{pintelas2025mobilenet}. Unlike classical training overfitting (TO), which results from excessive memorization of training data, VO arises when the model selection process becomes overly adapted to the specific characteristics, noise, or biases of the validation set.

The root cause of VO lies in the excessive reliance on validation feedback for model evaluation, selection, and early stopping. When a model is evaluated repeatedly against the same validation set, especially in low-data or imbalanced scenarios \cite{gaudreault2024empirical}, the probability of observing artificially high validation scores by chance increases. As a result, models may be chosen not for their true generalization capabilities but for coincidental alignment with the validation data's peculiarities.

Let $\mathcal{D}_{T}$, $\mathcal{D}_{V}$, and $\mathcal{D}_{E}$ denote the training, validation, and evaluation datasets, respectively, with $\mathcal{D}_{V}$ serving as a proxy to estimate generalization performance. In ideal conditions, performance on $\mathcal{D}_{V}$ should correlate strongly with performance on $\mathcal{D}_{E}$. However, under repeated evaluations and selection pressure, the following degradation can occur:
\vspace*{-0.05cm}
$$
\text{Perf}(\mathcal{D}_{T}) \gg \text{Perf}(\mathcal{D}_{V}) \approx \text{Perf}(\mathcal{D}_{E}) \quad \text{(TO)},
$$
and 
$$
\text{Perf}(\mathcal{D}_{V}) \gg \text{Perf}(\mathcal{D}_{E}) \quad \text{(VO)}.
$$
In training overfitting \cite{aburass2024quantifying}, the model performs well on training data but poorly on both validation and test data. In contrast, under validation overfitting, the model appears optimal on the validation set but fails dramatically on the test set, giving a false sense of reliability during development \cite{zhao2024breaking}.

VO is particularly dangerous because it is often invisible until final testing or deployment phases, where operational data significantly diverges from validation samples. This risk is exacerbated when: the validation set is small, biased, or unrepresentative; the number of validation checks during training or model search is high; model selection strategies aggressively optimize for validation metrics. 
Thus, while validation monitoring remains a fundamental technique to avoid training overfitting, over-dependence on validation evaluation without safeguards introduces a critical failure mode via VO.

In this context, minimizing the number of validation checks, promoting diversity in model generation, and decoupling model construction from heavy validation optimization, emerge as promising strategies to mitigate validation overfitting risk. These insights directly motivate the design principles of the proposed GeNeX framework.

\subsection{Formalizing Validation Overfitting}

We formalize validation overfitting (VO) along two perspectives: (i) the \textit{modeling perspective}, focusing on the pipeline of model generation and ensemble construction; and (ii) the \textit{dataset perspective}, where VO arises intrinsically due to underlying distributional shifts between development and evaluation data. 

In the modeling view, VO emerges during both the generation of base models (often involving heavy validation monitoring) and the strategy used to select ensemble members. This includes widely-used sequential selection methods that incrementally select ensemble members based on validation performance gains; a common approach in both practice and competitive settings \cite{sarmah2024ensemble}. On the dataset side, VO risk increases when the validation set is not representative of the testing distribution, especially in real-world or competition-style environments with intrinsic domain or acquisition shifts~\cite{zhao2024breaking,xu2024cbdmoe}.

\subsubsection{Model-Induced VO Risk}\label{Model-Induced VO Risk}

To promote diversity, a pool of \( N \) models is typically generated via $N$ independent training runs, each performed under varying configurations (e.g., random seeds, data augmentations, dropout schemes). Each model is trained using a standard train-validation loop, during which validation performance is monitored across $q$ checkpoints (e.g., per epoch), and the best-performing checkpoint is retained. 
This process results in a total of \( N \times q \) validation queries during the model generation phase alone. This increases the likelihood that some models in the candidate pool $\mathcal{M} = \{ f_{\theta^{(1)}}, f_{\theta^{(2)}}, \dots, f_{\theta^{(N)}} \}$ will exploit dataset-specific noise or statistical flukes, resulting in overly optimistic validation scores.

Let $L_{i,t}$ denote the observed validation loss for run $i$ at checkpoint $t$, 
and decompose it as
$L_{i,t} = \nu_{i,t} + Y_{i,t}$,
where $\nu_{i,t} \coloneqq \mathbb{E}[L_{i,t}]$ is the expected validation loss 
with respect to the underlying data distribution, 
and the zero-mean term $Y_{i,t}$ captures the random fluctuation due to finite-sample effects, 
so that $\mathbb{E}[Y_{i,t}] = 0$.
Define the best (true) population validation loss as
\(
\nu_\star \coloneqq \min_{i \le N,\, t \le q} \nu_{i,t}.
\)
Then
\[
\begin{aligned}
\mathbb{E}\!\left[\min_{i\le N,\,t\le q} L_{i,t}\right]
&= \mathbb{E}\!\left[\min_{i,t}\big(\nu_{i,t}+Y_{i,t}\big)\right] \\[0.3em]
&\ge \nu_\star + \mathbb{E}\!\left[\min_{i,t} Y_{i,t}\right]
= \nu_\star - \mathbb{E}\!\left[\max_{i,t}(-Y_{i,t})\right].
\end{aligned}
\]
If the centered noise $Y_{i,t}$ is sub-Gaussian with variance proxy $v$ (scale $s_m^2=v$) on $\mathcal{D}_V$, using the inequality for the expectation of a maximum \cite[Sec.~2.5, p.~31]{blm2013},
we have $\mathbb{E}\max_{j\le M} Z_j \le \sqrt{2v\log M}$ for $M = Nq$.
Applying this to $Z_j\equiv -Y_{i,t}$ yields
\[
\mathbb{E}\!\left[\min_{i\le N,\,t\le q} L_{i,t}\right]
\;\ge\;
\nu_\star \;-\; s_m\,\sqrt{2\log(Nq)}.
\]
More generally, if $Y_{i,t}$ is sub-gamma with parameters $(v,c)$, \cite[Sec.~2.5, Cor.~2.6]{blm2013} gives
$\mathbb{E}\max_{j\le M} Z_j \le \sqrt{2v\log M}+c\,\log M$. Hence,
\[
\mathbb{E}\!\left[\min_{i,t} L_{i,t}\right]
\;\ge\;
\nu_\star \;-\; \Big(\sqrt{2v\log(Nq)} + c\,\log(Nq)\Big).
\]
The above derivation quantifies the expected optimism 
$\nu_\star - \mathbb{E}[\min_{i,t} L_{i,t}]$, 
capturing how much the best observed validation loss 
appears better than its true expected value. 
It grows at least on the order of 
$\mathcal{O}(\sqrt{\log(Nq)})$, indicating that enlarging 
the search space (through more runs or checkpoints) inevitably 
amplifies VO.

The risk of VO is then further influenced during ensemble construction. 
Let $L_V(\mathcal{E})$ denote the validation loss of an ensemble $\mathcal{E}$ 
evaluated on $\mathcal{D}_V$. 
Suppose the ensemble is built incrementally by selecting, at iteration $k$,
\[
\mathcal{E}_k = \mathcal{E}_{k-1} \cup \{ f_{\theta^{(k)}} \},
\quad
f_{\theta^{(k)}} \in 
\arg\min_{f \in \mathcal{M} \setminus \mathcal{E}_{k-1}}
L_V\!\big(\mathcal{E}_{k-1} \cup \{f\}\big),
\]
that is, by greedily adding the candidate that yields the largest 
reduction in validation loss.
Each selection step thus introduces an additional layer of 
validation-driven optimization, effectively replacing $\mathcal{M}$ 
with a sequence of model subsets whose composition depends on 
repeated access to $\mathcal{D}_V$.
Consequently, the ensemble selection trajectory 
$\{\mathcal{E}_k\}_{k=1}^K$ can accumulate multiple opportunities for 
validation overfitting, since each incremental decision exploits random 
fluctuations in $L_V$ in a manner analogous to the 
$\min_{i,t} L_{i,t}$ effect analyzed above.
In expectation, this induces a compounded optimism bias that scales with 
both the ensemble size $K$ and the effective search complexity of the 
candidate pool, thereby amplifying the gap between validation and test 
performance estimates.

To summarize, the overall model-induced VO risk is governed by the following three factors:

\begin{itemize}
    \item The number of validation checkpoints $q$ per training run, which amplifies exposure to the validation set and increases the likelihood of overfit checkpoints.
    \item The number of models $N$ in the candidate pool, which controls the chance of including validation outliers.
    \item The final ensemble size $K$, which governs how many potentially biased models are aggregated.
\end{itemize}

\vspace{-\parskip}
\noindent
Together, these factors suggest that VO risk increases with the overall validation query budget ($N \times q$), and with the aggressiveness of validation-driven ensemble selection ($K$).

\subsubsection{Dataset-Induced VO Risk}

A small and unrepresentative validation set can lead to VO, especially if model-developing prioritize validation performance as described before. This can intrinsically arise from inherent distributional differences between the validation and test sets, referred also as ``distributional-shift'', which is commonly seen in competition-driven or real-world environments \cite{zhao2024breaking,xu2024cbdmoe}. 

To formalize this, we analyze the test-time generalization risk in the presence of a class prior shift between validation and testing distributions. Let \[
\pi_{V} = (\pi_{V,1}, \dots, \pi_{V,C}) \quad
\text{and} \quad
\pi_{E} = (\pi_{E,1}, \dots, \pi_{E,C}),
\]
denote the class prior distributions in the validation and test sets, respectively, where $C$ is the number of classes.

Following~\cite{zhao2024breaking}, let $R_{V}(f)$ and $R_{E}(f)$ be the expected risk of a classifier $f$ on validation and test data:
\[
R_{V}(f) = \sum_{i=1}^C \pi_{V,i} \cdot \mathbb{E}_{x \sim \mathcal{D}_{V}(x|y=i)}[\ell(f(x), i)],
\]
\[
R_{E}(f) = \sum_{i=1}^C \pi_{E,i} \cdot \mathbb{E}_{x \sim \mathcal{D}_{E}(x|y=i)}[\ell(f(x), i)].
\]
Assuming conditional distributions $\mathcal{D}(x|y)$ are approximately stable (i.e., label shift setting), the test risk can be decomposed as:
\[
R_{E}(f) = R_{V}(f) + \sum_{i=1}^C \left( \pi_{E,i} - \pi_{V,i} \right) \cdot \mathbb{E}_{x \sim \mathcal{D}(x|y=i)}[\ell(f(x), i)].
\]
Bounding the per-class expected loss by a constant $M$, we obtain:
\[
|R_{E}(f) - R_{V}(f)| \leq M \cdot \sum_{i=1}^C |\pi_{E,i} - \pi_{V,i}| = 2M \cdot \delta_{\text{TV}}(\pi_{V}, \pi_{E}),
\]
where $\delta_{\text{TV}}$ is the total variation distance between class priors.

This bound shows that the generalization risk grows proportionally with the magnitude of class prior shift. In other words, the greater the distributional shift between the validation and test sets, the higher the risk of validation overfitting becomes.

To capture broader distributional shifts, we employ the Jensen-Shannon Divergence (JSD) \cite{englesson2021generalized} between the model's predictive output distributions on the validation and test sets. Let $P$ and $Q$ denote the softmax output distributions of a model $f$ on validation and test sets, respectively. The JSD is defined as:
\vspace*{-0.05cm}
\[
\text{JSD}(P \parallel Q) = \frac{1}{2} D_{\text{KL}}(P \parallel M) + \frac{1}{2} D_{\text{KL}}(Q \parallel M),
\]
where \( M = \frac{1}{2}(P + Q) \) and \( D_{\text{KL}}(P \parallel Q) \) is the Kullback-Leibler divergence:
\[
D_{\text{KL}}(P \parallel Q) = \sum_{i} P(i) \log \frac{P(i)}{Q(i)}.
\]
JSD is symmetric and bounded in $[0, 1]$, making it suitable for quantifying validation-to-test shift. 
A high JSD implies strong distributional shift between the validation and test sets; thus, the risk of validation overfitting increases and the performance on $\mathcal{D}_{V}$ becomes an unreliable proxy for performance on $\mathcal{D}_{E}$.

This justifies the use of JSD in our VO-aware evaluation protocol. Specifically, we use JSD in two ways: (i) to construct validation/test splits with controlled shifts for benchmarking, and (ii) to assess VO robustness post-hoc. High JSD coupled with small VO gaps and high test accuracy indicates a method’s ability to generalize across significant shifts, a key property of VO-resilient pipelines.

\subsection{Implications and Summary}

Based on the previous analysis, we can conclude that validation overfitting is tightly linked to:
\begin{itemize}
    \item Ensemble-modeling perspective:
        \begin{itemize}
            \item the generation strategy followed to train individual base models, which is highly relying to the total number (\( N \)) of generated models,
            \item and the selection strategy followed to select the members (\( K \)) of the final ensemble predictor.
            % to us link with number of final ensembe (number of clusters -> number of prototypes (here is why it is also important this prototype based fusion that decreases also K while promotes diversity by representing each cluster .....
        \end{itemize}
        Larger \( N \) increases the chance of selecting validation outliers, while larger \( K \) compounds noise accumulation during ensemble construction.

    \item Dataset perspective: which relates with the inherent distributional-shift magnitude between training/validation and testing sets that further amplifies the risk of VO during model developing.
\end{itemize}

Ensemble pipelines that heavily rely on validation feedback during model generation and selection are particularly vulnerable to these effects. Consequently, to mitigate validation overfitting in ensemble pipelines, it is crucial to: minimize the number of validation queries during base models generation, promote intrinsic diversity among models without heavy validation reliance and reduce final ensemble size while maximizing representational diversity. These insights form the theoretical motivation behind the design of the proposed GeNeX framework, which aims to minimize validation reliance and promote representational diversity, to ensure robust generalization and minimize VO risk.

Moreover, to ensure rigorous benchmarking of VO robustness, evaluation protocols must explicitly simulate real-world shifts. The use of the proposed JSD-guided partitioning allows for controllable distributional divergence between training/validation and testing sets, enabling researchers to create VO-prone scenarios by design in any case-study dataset.

Beyond split construction, in our framework, JSD is used as a post-hoc indicator to assess dataset difficulty and VO risk. A model achieving high test accuracy and low VO gap under a high-JSD split provides compelling evidence of robustness and generalization. Conversely, performance in low-JSD scenarios (i.e., in-distribution) carries less evidential weight of model's robustness against VO, as such conditions are inherently easier.

\section{Related work}

Having established the theoretical foundations and challenges of validation overfitting, we now examine existing research across three key areas: (a) standard ensemble learning strategies for robustification, (b) evolutionary and genetic-based approaches for promoting ensemble diversity, and (c) expert-aware and validation-resilient ensemble formulations.

\paragraph{Classical Ensemble Learning}
Ensemble methods such as bagging, boosting, and snapshot ensembling are long-established techniques for improving model generalization and reducing variance \cite{snapshot_ensemble,swa_ensemble}. These approaches promote diversity through data resampling, stochastic training schedules, or architectural variation. While effective against training overfitting, these strategies remain susceptible to validation overfitting, especially when model selection or ensemble formation relies heavily on validation score improvements \cite{pintelas2025mobilenet}. 

Widely used incremental ensemble methods \cite{sarmah2024ensemble} progressively select models based on validation gains, aiming to iteratively improve validation accuracy. However, such strategies risk accumulating noise from validation fluctuations, particularly in low-data or high-variance settings, leading to validation overfitting \cite{pintelas2025textnex}.

\paragraph{Genetic and Evolutionary Ensemble Construction}
To mitigate validation reliance and promote model diversity, population-based and genetic approaches have gained traction. Recent works have proposed hybrid frameworks combining deep learning with genetic algorithms for medical image analysis \cite{kaya2024novel,dar2024deep}, or for instance selection in ensembles \cite{xu2024genetic}. These methods generate new models through crossover and mutation operations in weight or feature space, promoting exploration of under-represented regions in the model space. While effective in enhancing diversity and bypassing over-optimized local minima, they often stop short of formal VO mitigation during ensemble construction.

\paragraph{Expert-Aware and VO-Resilient Ensemble Formulations}
A newer line of research focuses on explicitly managing behavioral diversity in ensembles through expert modeling and controlled generalization.
In a recent work \cite{zhao2024breaking}, the authors propose a controllable expert generation framework using hypernetworks to inject targeted diversity across long-tailed data regimes, demonstrating robustness under domain distribution shifts. Similarly, CBDMoE~\cite{xu2024cbdmoe} introduces a consistent-but-diverse mixture of experts architecture to improve domain generalization without sacrificing model consistency. These works highlight the importance of crafting ensembles from complementary sub-models while balancing robustness and diversity.
Complementary to the above, heterogeneity-driven expert ensemble methods \cite{pintelas2025mobilenet,pintelas2025textnex} have recently emerged. These methods cluster models by prediction behavior and select one representative expert per cluster using criteria like validation score or centroid proximity. While these strategies improve model diversity and mitigate VO to some extent, they remain vulnerable, since the initial model generation relies on validation-based monitoring, increasing the risk of noisy models to infiltrate clusters, compromising the final ensemble's integrity.

\paragraph{Proposed Approach} In contrast to all previous works, the proposed GeNeX framework directly addresses validation overfitting at both model generation and ensemble construction/selection stages. During model generation, GeNeX entirely avoids the use of validation feedback by employing a dual-path strategy that combines gradient-based supervised training with genetic-based model evolution. This design is grounded in the hypothesis that generating offspring networks via crossover of gradient-trained parent models acts as a weight-level regeneration mechanism—mitigating overfitting while promoting structural diversity. As a result, GeNeX forms a candidate model pool that is both diverse and robust, without introducing validation-overfitted candidates.

At the ensemble construction stage, instead of selecting a single representative expert per cluster, GeNeX elects multiple diverse experts using complementary criteria and fuses them via weight-level averaging into robust prototype models. This prototype-based construction ensures representational richness and behavioral complementarity within the final ensemble. Overall, GeNeX achieves strong generalization with minimal reliance on validation signals, offering enhanced robustness in both limited-data and distributionally-shifted scenarios.

\section{GeNeX framework}

In this section, we present the core design principles and components of the proposed GeNeX framework, detailing how its dual-module architecture aims to address validation overfitting by rethinking both model generation and ensemble construction. The implementation code is made available at the following link\footnote{\href{https://github.com/EmmanuelPintelas/Genetic-Network-eXperts-GeNeX-}{GeNeX implementation link}}.

\subsection{Overview of the GeNeX Framework}

The GeNeX framework is designed to systematically address the challenge of validation overfitting (VO) by minimizing the dependency on validation feedback during model generation and ensemble construction. It consists of two main modules: Genetic-based Ensemble constructor (GenE) and Prototype-based Network eXperts selector (ProtoNeX).

The first module, GenE, aims to create a diverse and robust population of base networks through a hybrid training strategy. In each iteration, networks are generated via two complementary paths: (i) supervised training with diverse configurations, under zero validation dependency, and (ii) genetic weight fusion, where model weights are combined and mutated to generate new offspring networks. This genetic process aims to refresh potentially overfitted networks by encouraging exploration of the model space beyond gradient-based optimization, promoting diversity and robustness.

The second module, ProtoNeX, focuses on building a complementary ensemble from the generated network pool. Networks are clustered based on their predictive behavior, and multiple eXpert models are elected within each cluster based on diverse criteria, such as validation performance, robustness to perturbations, representativeness, and anomalous behavior. The selected experts within each cluster are then fused at the weight level to create a prototype model that encapsulates the cluster's representational diversity. Finally, all prototype models are combined into a final ensemble, with the ensemble weights optimized via Sequential Quadratic Programming \cite{virtanen2020scipy} to maximize output-level synergy.

\begin{figure*}[!ht]
\centering
%\hspace*{-0.7cm} 
\includegraphics[width=0.90\linewidth]{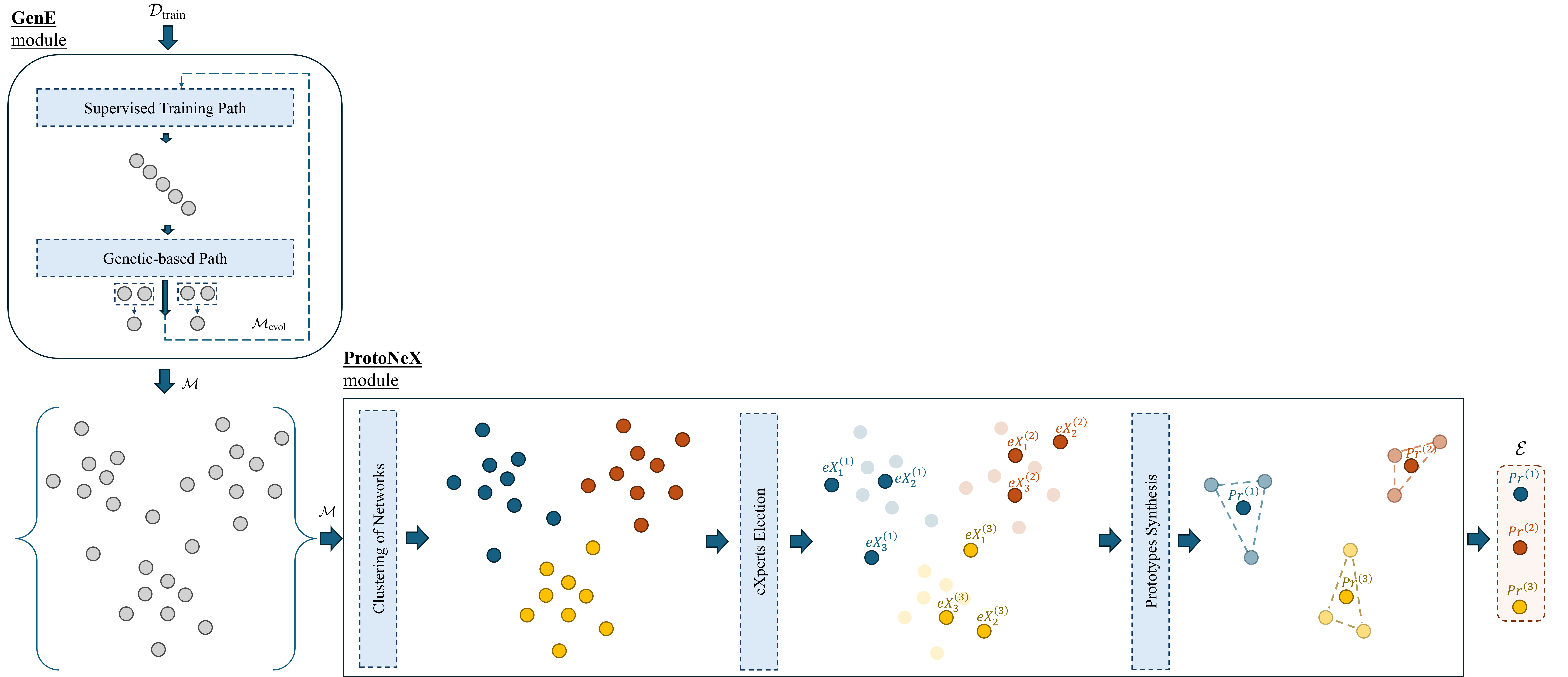}
% \caption{Overview of the GeNeX framework. It consists of the GenE module, generating a diverse pool of networks $\mathcal{M}$ via supervised training and genetic crossover, and the ProtoNeX module, which performs behavior-based clustering, selects diverse experts, and fuses them into prototypes to form the final ensemble $\mathcal{E}$. This design minimizes validation reliance and promotes complementary representation across the model space, addressing validation overfitting while enhancing generalization.}
\caption{GeNeX overview. GenE generates a diverse pool $\mathcal{M}$ without validation monitoring via short supervised training and genetic crossover/mutation, encouraging broad weight-space exploration and limiting early validation dependence. ProtoNeX clusters models in behavior (prediction) space, elects complementary experts via multi-criteria selection, and fuses them into $K$ compact prototypes. Instead of clustering data and training separate models per cluster, ProtoNeX clusters the models themselves and uses prototype fusion, promoting complementarity and behavioral diversity across the model space to enhance generalization.}

\label{genex_framework}
\end{figure*}

An abstract overview of the GeNeX pipeline is illustrated in Figure~\ref{genex_framework}, showcasing the main phases and the flow of model generation, clustering, expert selection, and prototype fusion. Through this multi-phase design, GeNeX aims to generate validation-robust ensembles that maintain high generalization performance while systematically reducing the risk of validation overfitting.

\subsubsection{GenE Module: Genetic-based Ensemble generator}

The first module of the GeNeX framework, referred to as GenE (Genetic-based Ensemble generator), aims to generate a diverse and robust pool of base networks $\mathcal{M}$ with minimal reliance on validation feedback. To achieve this, GenE integrates two complementary model generation paths: a \textit{supervised} and \textit{genetic-based}.

\paragraph{Supervised Training Path}

In the supervised training path, a set of base networks are trained conventionally using labeled data and gradient-based optimization. To promote diversity within the population, various configurations are randomized across models, including different random seeds, data augmentations, dropout rates, learning rate schedules, and initialization schemes. This step ensures that even purely supervised models explore distinct regions of the model space.

Unlike standard deep learning pipelines that rely on validation monitoring, this phase deliberately avoids selecting best-validation checkpoints; models are trained for a fixed, small number of epochs (1--3).

\paragraph{Genetic-based Path}

Parallel to supervised training, GenE introduces a genetic-based model generation mechanism inspired by evolutionary algorithms. In this path, random pairs of independent-trained parent networks are selected from the existing pool, and new child networks are generated via weight-level fusion and mutation. The genetic operations include:

Weight Fusion (Crossover): Given two parent networks, their corresponding weight tensors are combined through element-wise operations (e.g., averaging, interpolation, or randomized mask-based selection) to generate the child's initial weights. This operation allows knowledge from both parents to be inherited and mixed, potentially creating new representational patterns without relying on gradient flows. 

Mutation: After fusion, controlled Gaussian noise is injected into random layers or parameters of the child network. Mutation encourages exploration of the weight space, introducing new variations that may lead to better generalization.

Children networks generated through this genetic process are lightly fine-tuned (last task-specific layer) on training data for a small number of epochs, further maturing and adapting their representations without validation supervision.

Through iterative supervised training and genetic operations, the model population gradually expands, encompassing a wide variety of architectures, training trajectories, and weight configurations. Crucially, because model generation relies primarily on intrinsic diversity mechanisms rather than validation-based selection, the resulting population is less vulnerable to validation overfitting.

\subsubsection{ProtoNeX Module: Prototype-based Network eXperts selector}

After generating a diverse pool of base networks through the GenE module, the ProtoNeX module aims to construct a robust and complementary ensemble predictor $\mathcal{E}$, by systematically selecting and synthesizing representative expert models.

% \vspace{-\parskip}
% \textit{a) Behavioral Clustering of Networks:}
\paragraph{Behavioral Clustering of Networks}
To organize the generated network pool, we first cluster the models based on their prediction behaviors. Specifically, models are embedded into a feature space derived from their output vectors on a shared validation subset, capturing their functional similarity. Clustering is then performed via Gaussian Mixture Models \cite{zhang2021gaussian}, grouping models into subpopulations that exhibit similar predictive patterns.
This clustering step ensures that models specializing in different aspects of the data distribution are separated, facilitating complementary expert selection in subsequent steps.

% \vspace{-\parskip}
% \textit{b) Expert Election Within Clusters:}
\paragraph{Expert Election Within Clusters}
Within each cluster, multiple expert networks are elected based on diverse and orthogonal criteria, ensuring that various traits of model quality and behavior are captured:
the top validation performer, the most robust under perturbations, the representative closest to the cluster centroid, the intra-anomalous expert farthest from it (capturing rare behaviors), and the outer-anomalous expert being most dissimilar to other clusters.
% \begin{itemize} \item Top-Validation Performer: The model with the highest validation score. \item Most Robust Model: The model exhibiting the least performance degradation under input perturbations or noise. \item Cluster Representative: The model closest to the cluster centroid in prediction space. \item Intra-Anomalous Expert: The model furthest from the cluster centroid, capturing potential novel or rare behaviors within the cluster. \item Outer-Anomalous Expert: The model most dissimilar to models from other clusters, promoting inter-cluster diversity. \end{itemize}
By electing multiple experts with different selection criteria, ProtoNeX ensures that each cluster’s representational space is thoroughly and reliably captured, reducing the risk of noisy or biased selections.

% \vspace{-\parskip}
% \textit{c) Prototype Synthesis via Weight Fusion:}
\vspace{-\parskip}
\paragraph{Prototype Synthesis via Weight Fusion}
%\vspace{-0.5em}
Rather than selecting a single expert per cluster, ProtoNeX synthesizes a prototype model by fusing the internal weights of the elected experts. Weight fusion is performed via layer-wise averaging, aligning corresponding weights across models.
This prototype model encapsulates the collective knowledge, robustness, and diversity of its cluster, offering a lightweight yet expressive representation without inflating model size.
Finally, each prototype model is lightly fine-tuned (last task-specific layer) on training data, without validation supervision, to adapt their fused representations.

% \vspace{-\parskip}
% \textit{d) Ensemble Weight Optimization:}
\vspace{-\parskip}
\paragraph{Ensemble Weight Optimization}

%\vspace{-1em}
Once all cluster prototypes are generated, they are combined into the final ensemble via Sequential Quadratic Programming (SQP), yielding a convex combination of prototype predictions. Convex weighting can outperform simple averaging, especially when base models differ in reliability~\cite{choi2020combining}. At the same time, restricting the combiner to a simplex-constrained convex combination gives a low-capacity model whose generalization properties are better controlled than those of unconstrained linear or more complex nonlinear stacking schemes~\cite{ijcai2020p429}. In our case the combiner introduces only a small number of simplex-constrained parameters (one weight per prototype), which has substantially lower capacity than typical nonlinear stacking models (e.g., neural-network meta-learners or higher-order polynomial combinations of base predictions), reducing the risk of overfitting the validation set.

\vspace{-\parskip}
Through this structured ensemble construction, ProtoNeX builds a final model that is both robust to VO and capable of strong generalization across diverse real-world tasks.

\subsection{Theoretical View of GeNeX Methodology}

The proposed GeNeX framework is designed to mitigate validation overfitting by enforcing diversity, complementarity, and representational robustness at multiple stages of model generation and ensemble construction.

\subsubsection{Genetic process as Diversity-Inducing and Overfitting-Resilient Path}\label{study2}

Let the model population be initialized as a set of $N$ networks trained via supervised learning:
\(
\mathcal{M} = \{ f_{\theta^{(i)}}^{\text{grad}} \}_{i=1}^{N},
\)
where each model is trained with randomized configurations (e.g., seeds, augmentations) under gradient-based optimization. 
GenE avoids checkpoint-based validation monitoring during model generation so as to minimize validation overfitting. Under the same maximal-inequality view used in Section~\ref{Model-Induced VO Risk}, the expected optimism from selecting among $M$ validation queries scales as $\mathcal{O}\!\big(\sqrt{\log M}\big)$; standard pipelines induce $M{=}Nq$ via $N$ runs and $q$ checkpoints, whereas GenE’s generation stage contributes no such querying and any later selection is independent of $q$ (at most $\mathcal{O}(N)$), effectively removing the $\log q$ term. 
% Incorporating validation monitoring here would result in an $N \times q$ total validation-query dependency (as discussed in Section~\ref{vo_definition}), substantially increasing the risk of injecting potentially ``rotten roots'' (i.e., validation-overfitted candidates) into the subsequent ensemble construction pipeline.
However, via such an approach, classic training-overfitting remains a concern; here comes our hypothesis that a genetic-based process could refresh the population to mitigate training and validation overfitting while promoting diversity.
Let $f_{\theta^{(j)}}^{\text{gen}}$ denote a child model generated by fusing weights of two parent models from $\mathcal{M}$:
\(
\theta^{(j)} = \alpha \theta^{(a)} + (1 - \alpha) \theta^{(b)} + \epsilon,
\)
where $\alpha \in [0, 1]$ controls crossover balance and $\epsilon \sim \mathcal{N}(0, \sigma^2)$ models mutation noise.
Unlike gradient descent, GenE’s operators act as geometric regularizers in weight space. Crossover and small-variance mutation moves offspring away from sharp parent minima toward broader, flatter valleys in the loss landscape. This mechanism is aligned with evidence that weight averaging and weight-space interpolation yield flatter optima and better generalization, thereby narrowing training/validation–test gaps~\cite{swa_ensemble,garipov2018loss,wortsman2022model}.

\textbf{Claim:} Given two overfit parents $f_{\theta^{(a)}}, f_{\theta^{(b)}}$, their fusion under random crossover and mutation produces a model $\mathbb{E}[f_{\theta^{(j)}}^{\text{gen}}]$ that lies in a region less likely to inherit overfit features directly, acting as a diversity-preserving corrective path.

\subsubsection{Behavioral Clustering as Complementarity Enforcer}

Let $\mathcal{M}$ be the full population of trained and genetically evolved models. For each model $f_{\theta}$, we extract its prediction vector $z_{\theta} \in \mathbb{R}^d$ (e.g., concatenated softmax outputs on a shared validation subset). 
To identify groups of behaviorally similar models, we cluster these prediction vectors using \emph{Gaussian Mixture Models (GMM)}. Compared to hard-partitioning methods such as $k$-means, GMM yields a probabilistic, soft clustering and allows anisotropic, overlapping clusters in high-dimensional spaces, which is well suited for modeling correlated prediction patterns of neural networks~\cite{patel2020clustering,zhang2021gaussian,murphy2022probabilistic}. This choice therefore provides a more expressive representation of behavioral similarity than common spherical-cluster or distance-only alternatives~\cite{murphy2022probabilistic}. In this context, each cluster is modeled as a multivariate Gaussian distribution with its own mean $\mu_k$ and covariance matrix $\Sigma_k$. The assignment of a model to a cluster is based on the component that gives the highest likelihood under the Gaussian probability density function:
\(
\text{Cluster}(z_{\theta}) = \arg\max_{k \in \{1,\dots,K\}} \mathcal{N}(z_{\theta}; \mu_k, \Sigma_k),
\)
where $\mathcal{N}(z_\theta; \mu_k, \Sigma_k)$ denotes the multivariate Gaussian density function:
\(
\frac{1}{(2\pi)^{d/2} |\Sigma_k|^{1/2}} \exp\left( -\frac{1}{2}(z - \mu_k)^T \Sigma_k^{-1} (z - \mu_k) \right),
\)
and $d$ is the dimensionality of the prediction vectors.

Each cluster $\mathcal{C}_k$ groups models with similar predictive characteristics, allowing us to assign distinct semantic roles or ``expert behaviors'' to different subpopulations. This enables:
Intra-cluster specialization; within each $\mathcal{C}_k$, expert models may capture distinct patterns specific to a data subspace.
Inter-cluster complementarity; across $\{ \mathcal{C}_1, \dots, \mathcal{C}_K \}$, clusters span diverse predictive behaviors, ensuring that the final ensemble captures a broader function space.
This clustering step plays a critical role in organizing the model pool into behaviorally distinct groups, enabling the following expert election and prototype fusion stages to focus on complementary model traits from each cluster.

\subsubsection{Prototype-Based Weight Fusion as Representational Aggregation}

Within each cluster $\mathcal{C}_k$, a set of expert models $\{f_{\theta^{(k,1)}}, \dots, f_{\theta^{(k,m)}}\}$ is selected based on orthogonal criteria (e.g., top-validation, robust to noise, representative, anomalous). These experts are fused into a prototype model via layer-wise averaging:
\vspace*{-0.05cm}
\(
\theta^{(k)} = \frac{1}{m} \sum_{j=1}^{m} \theta^{(k,j)}.
\)
This operation aggregates both the dominant and diverse representational traits within each cluster. We define the prototype network $f_{\theta^{(k)}}$ as:
\vspace*{-0.05cm}
\(
f_{\theta^{(k)}} = \text{Prototype}(\mathcal{C}_k) = \arg\min_{\theta} \sum_{j=1}^{m} \| \theta - \theta^{(k,j)} \|^2,
\)
i.e., a centroid in weight space that minimizes intra-cluster variance.

\textbf{Claim:} The prototype network approximates the central representation of its cluster while preserving performance-relevant properties from its expert constituents. 

Thus, combining $K$ such prototypes, we form an ensemble:
\(
\mathcal{E} = \sum_{k=1}^{K} w_k f_{\theta^{(k)}},
\)
with combination weights $w_k$ optimized via SQP.

\subsection{Algorithmic View of GeNeX framework}

The GeNeX framework is composed of two sequential modules: (i) GenE, which generates a pool $\mathcal{M}$ of $N$ diverse networks, and (ii) ProtoNeX, which transforms $\mathcal{M}$ into a robust ensemble $\mathcal{E}$ predictor.

\begin{algorithm}[!ht]
\small{
\caption{GenE Module: Genetic-based Ensemble \\Generation}
\label{alg:gene}
\textbf{Input:} Training set $\mathcal{D}_{T}$, base configuration set $\mathcal{C}$, total target models $N$, number of generations $G$, genetic offspring per generation $N_g$ \\
\textbf{Output:} Model pool $\mathcal{M}$ of size $N$

\begin{algorithmic}[1]
\vspace{0.3em}
\STATE \textbf{Initialization}
\STATE Initialize output model pool $\mathcal{M} \leftarrow \emptyset$
\STATE Set per-generation selection quota $s \leftarrow \frac{N}{G}$

\vspace{0.5em}
\STATE \textbf{Gradient-based Initialization (Generation 0)}
\STATE Initialize evolutionary model pool $\mathcal{M}_{\text{evol}} \leftarrow \emptyset$
\FOR{$i = 1$ to $N$}
    \STATE Sample configuration $\mathcal{C}_i \in \mathcal{C}$
    \STATE Train model $f_{\theta^{(i)}}$ on $\mathcal{D}_{T}$ (no validation monitoring)
    \STATE Add model to pool: $\mathcal{M}_{\text{evol}} \leftarrow \mathcal{M}_{\text{evol}} \cup \{f_{\theta^{(i)}}^{\text{grad}}\}$
\ENDFOR

\vspace{0.5em}
\STATE \textbf{Evolution Loop}
\FOR{$g = 1$ to $G$}

    \STATE \textit{// Genetic Child Generation and Light Fine-tuning}
    \STATE Initialize temporary set $\mathcal{M}_{\text{gen}} \leftarrow \emptyset$
    \FOR{$j = 1$ to $N_g$}
        \STATE Randomly select parents $f_{\theta^{(a)}}, f_{\theta^{(b)}} \in \mathcal{M}_{\text{evol}}$
        \STATE Generate child weights via crossover and mutation:
        \(
        \theta^{(j)} = \alpha \theta^{(a)} + (1 - \alpha) \theta^{(b)} + \epsilon,\quad \epsilon \sim \mathcal{N}(0, \sigma^2)
        \)
        \STATE Lightly fine-tune $f_{\theta^{(j)}}$ on $\mathcal{D}_{T}$ (no validation monitoring)
        \hspace*{-0.2cm}
        \STATE Add to temporary pool: $\mathcal{M}_{\text{gen}} \leftarrow \mathcal{M}_{\text{gen}} \cup \{f_{\theta^{(j)}}^{\text{gen}}\}$
    \ENDFOR

    % \STATE Evaluate $\mathcal{M}_{\text{gen}}$ on training loss
    \STATE Select randomly $s$ models: $\mathcal{M}_{\text{sel}} \leftarrow \text{rand-}s(\mathcal{M}_{\text{gen}})$
    \STATE Update output pool: $\mathcal{M} \leftarrow \mathcal{M} \cup \mathcal{M}_{\text{sel}}$
    \STATE \textit{// Gradient-based Training on Remaining Candidates}
    \STATE Define $\mathcal{M}_{\text{rest}} = \mathcal{M}_{\text{gen}} \setminus \mathcal{M}_{\text{sel}}$
    \STATE Add $N_{\text{new}}$ fresh random initializations to $\mathcal{M}_{\text{rest}}$
    \STATE Discard previous generation: $\mathcal{M}_{\text{evol}} \leftarrow \emptyset$
    \FOR{each model $f_\theta \in \mathcal{M}_{\text{rest}}$}
        \STATE Sample new configuration $\mathcal{C}_k \in \mathcal{C}$
        \STATE Train on $\mathcal{D}_{T}$ (no validation monitoring)
        \STATE Update evolutionary pool: $\mathcal{M}_{\text{evol}} \leftarrow \mathcal{M}_{\text{evol}} \cup \{f_\theta^{\text{grad}}\}$
    \ENDFOR
\ENDFOR
\end{algorithmic}
}

\end{algorithm}

The \textit{GenE} module (Algorithm~\ref{alg:gene}) constructs a validation-agnostic ensemble model pool through an iterative evolutionary process designed to maximize diversity and minimize overfitting. It begins with a supervised initialization phase (Steps 5--10), where $N$ models are trained independently using randomly sampled configurations from the configuration space $\mathcal{C}$. Each model is trained for a few epochs (e.g. 1–3) on the training set $\mathcal{D}_{T}$, without any validation monitoring or checkpointing, forming the initial evolutionary pool $\mathcal{M}_{\text{evol}}$. The core of GenE is a multi-generational loop (Steps 12--32), in which two complementary paths are executed:

Genetic Path (Steps 14--20): A set of $N_g$ offspring models is generated via random pairwise weight fusion of existing models in $\mathcal{M}_{\text{evol}}$. 
For each pair of parents $f_{\theta^{(a)}}, f_{\theta^{(b)}}$, a child is formed by interpolating their weights using a convex combination with a crossover rate $\alpha$ and an additive Gaussian mutation term (\(\theta^{(j)} = \alpha \theta^{(a)} + (1 - \alpha) \theta^{(b)} + \epsilon,\quad \epsilon \sim \mathcal{N}(0, \sigma^2)\)). In all experiments we fix $\alpha\!=\!0.5$ (balanced inheritance), which is qualitatively robust (very small $\alpha$ reduces novelty, while $\alpha\!\approx\!1$ drifts toward a single parent). The Gaussian mutation is applied layer-wise in convolutional/linear weight tensors (skipping buffers and batch-norm statistics) with probability $\rho$ per tensor. We fix $\rho{=}0.05$ and $\sigma{=}0.01$ across datasets; larger $(\rho,\sigma)$ increases exploration/diversity but can reduce stability, while smaller values under-explore. Mutation can be disabled by setting $\rho{=}0$. Finally, the resulting child models are lightly fine-tuned (output head) on $\mathcal{D}_{T}$ to stabilize them post-crossover-and-mutation, and temporarily stored in $\mathcal{M}_{\text{gen}}$.

\vspace{-\parskip}
Selection and Gradient Path (Steps 21--31): A random subset of $s$ models is selected from $\mathcal{M}_{\text{gen}}$ and added to the final output pool $\mathcal{M}$. The remaining $\mathcal{M}_{\text{rest}}$ children are forwarded to the supervised gradient path, where they are re-trained under new randomly sampled configurations. Additionally, $N_{\text{new}}$ freshly initialized models are introduced in each generation to further enhance diversity. These models, together with the non-selected children, form the next generation’s population $\mathcal{M}_{\text{evol}}$.
Over $G$ generations, this procedure yields a final output pool $\mathcal{M}$ of exactly $N$ models, each injected progressively across evolutionary stages. Importantly, with $s = N/G$ models added per generation, the final pool is composed of models from different evolutionary stages, promoting richer diversity in structural configurations and training trajectories. Crucially, the repeated injection of ``freshly'' initialized models throughout evolution prevents convergence to narrow regions of the weight space, further enhancing the robustness and complementarity of the resulting ensemble.

\vspace{-\parskip}
Algorithm~\ref{alg:protodistill} presents the core procedure of the \textit{ProtoNeX} module, which transforms a genetically-evolved diverse model pool into a compact and complementary ensemble by synthesizing prototype networks per behavioral cluster.
The algorithm accepts as input the full model pool \( \mathcal{M} \) (output from the preceding GenE stage), the training set \( \mathcal{D}_{T} \), the validation set \( \mathcal{D}_{V} \), and a user-specified number \( m \) of elected experts per cluster. The goal is to build a final ensemble \( \mathcal{E} \) consisting of $K$ prototype networks, one per behavioral cluster, each distilled from $m$ diverse experts selected under orthogonal criteria.

\begin{algorithm}[!ht]
\small{
\caption{ProtoNeX Module: Prototype-based Network eXperts Selection}
\label{alg:protodistill}
\textbf{Input:} Model pool $\mathcal{M}$, training set $\mathcal{D}_{T}$, validation set $\mathcal{D}_{V}$, number of elected experts per cluster $m$ \\
\textbf{Output:} Final ensemble model $\mathcal{E}$

\begin{algorithmic}[1]

\vspace{0.3em}
\STATE \textbf{Behavioral Clustering}
\STATE Inference each $f_{\theta} \in \mathcal{M}$ on $\mathcal{D}_{V}$ to collect prediction vectors
\STATE Apply UMAP/GMM-based clustering on prediction vectors $\Rightarrow$ partition into $K$ clusters $\{C_1, C_2, \dots, C_K\}$

\vspace{0.3em}
\STATE \textbf{Expert Election per Cluster}
\FOR{each cluster $C_k$}
    \STATE Elect $m$ experts using diverse criteria:
Top-Val Performer (max $\mathcal{A}_{V}$); Most Robust to perturbations; Cluster Representative (closest to centroid); Intra-Anomalous (furthest from centroid); Outer-Anomalous (most distant from other clusters)
    \STATE Denote selected experts as $E_k = \{f_{\theta_1}, \dots, f_{\theta_m}\}$
\ENDFOR

\vspace{0.3em}
\STATE \textbf{Prototype Synthesis via Weight Fusion}
\FOR{each $E_k$}
    \STATE Fuse weights of all $f_{\theta} \in E_k$ via layer-wise averaging.
    % \vspace*{-0.1cm}
    % \[
    % \theta^{\text{proto}_k} = \frac{1}{m} \sum_{i=1}^{m} \theta_i
    % \]
    %\vspace*{-0.2cm}
    \STATE Lightly fine-tune output head on $\mathcal{D}_{T}$ without val-monitoring.
    %to sharpen fused prototype
    \STATE Add prototype $f_{\theta^{\text{proto}_k}}$ to ensemble set: $\mathcal{E} \leftarrow \mathcal{E} \cup \{f_{\theta^{\text{proto}_k}}\}$
\ENDFOR

\vspace{0.3em}
\STATE \textbf{Ensemble Weight Optimization}
\STATE Use SQP to find optimal weights $\{w_1, \dots, w_K\}$. 
% that maximize:
% \[
% \mathcal{A}_{V}\left(\sum_{k=1}^K w_k f_{\theta^{\text{proto}_k}}(x)\right)
% \]

%\RETURN Final ensemble model $\mathcal{E}$

\end{algorithmic}
}
\end{algorithm}

\textit{Behavioral Clustering (Steps 2--3):}  
Each model \( f_\theta \in \mathcal{M} \) is fed on the validation set \( \mathcal{D}_{V} \) to extract its class-probability softmax vector, forming a high-dimensional prediction signature. These prediction vectors encode model behavior and are projected into a low-dimensional manifold using UMAP \cite{healy2024uniform}, capturing relative similarities among models based on their output distributions. Then, GMM is applied in the latent space to group behaviorally similar models into clusters. The number of clusters \( K \) is automatically determined using silhouette score maximization. Each cluster \( C_k \) ideally represents a functionally distinct sub-behavior (e.g., overconfident, conservative, robust-to-noise, etc.).

\textit{Expert Election per Cluster (Steps 5--8):}
Within each cluster \( C_k \), the algorithm elects $m$ complementary experts that exhibit diverse behavior, measured via five criteria. Top-Val Performer: model achieving the highest validation accuracy within the cluster; Most Robust: model whose predictions remain the most stable under test-time input perturbations; Cluster Representative: model closest to the GMM cluster centroid, indicating typical behavior; Intra-Anomalous: model most distant from the cluster centroid, capturing edge-case behavior; Outer-Anomalous: model most dissimilar to all other clusters, acting as an outlier from a global standpoint.
This expert election mechanism ensures that each cluster contributes not only its best-performing member but also diverse perspectives, including edge-case behaviors and stable predictors.

\textit{Prototype Synthesis via Weight Fusion (Steps 10--14):}  
For each cluster, the weights of the selected $m$ experts \( \{ \theta_1, \dots, \theta_m \} \) are averaged layer-wise to form a synthetic prototype \( \theta^{\text{proto}_k} \). This operation captures the common representational structure of the cluster and smooths out overfitted idiosyncrasies from individual experts. Each prototype is then lightly fine-tuned on the training set \( \mathcal{D}_{T} \) 
%using validation monitoring on \( \mathcal{D}_{V} \) 
to sharpen and adapt its output head. The refined prototype is added to the final ensemble set \( \mathcal{E} \).

\textit{Ensemble Weight Optimization (Step 16):}
To integrate the $K$ prototypes into a weighted ensemble, SQP is employed to optimize their scalar weights \( \{w_1, \dots, w_K\} \), maximizing the ensemble’s overall validation performance. This final step ensures that more contributive prototypes are weighted accordingly, resulting in a well-calibrated ensemble.

\textit{Compute, runtime, and scalability:}
GeNeX intentionally trades extra training-time exploration for low inference and memory overhead. In the GenE stage, the cost grows approximately linearly with the pool size $N$ and the number of generations $G$, and can parallelize across models because candidates are trained for only a few epochs and offspring are adjusted with light fine-tuning. ProtoNeX clusters by prediction signatures and fuses one prototype per cluster; the deployed predictor contains $K$ prototypes (typically $3$–$5$), so inference is $K$ forward passes (independent of $N$) and storage is $K$ checkpoints. Empirically, for $N\!\in\!\{5,500\}$, training time increases with $N$, while inference latency and memory remain essentially constant due to small $K$. Compared to classic ensembles that retain many models at test time, GeNeX concentrates computation in the offline training phase to gain behavioral diversity and minimize validation overfitting, while preserving fast and compact deployment.

\vspace{-1ex}
\section{Case Study Datasets}

To evaluate the effectiveness of the proposed GeNeX framework in mitigating validation overfitting and improving generalization, we conducted experiments across four diverse challenging real-world classification tasks.

\subsubsection{Skin Cancer classification}
Merging the datasets of the ISIC archive (thoroughly covering the period 2016–2024 \cite{rashid2025novel,hameed2024comprehensive}), a unified skin lesion image dataset was constructed, containing approximately 13,000 samples per class (\textit{benign}, \textit{malignant}). This task reflects a critical real-world application in dermatological diagnosis and early skin cancer detection.

\subsubsection{Deepfake detection}
A composite facial image dataset was constructed by merging three sources: the DeepFake Detection Challenge dataset~\cite{dolhansky2020deepfake}, DeepFakeMnist+~\cite{huang2021deepfake}, and EXPerts (a dataset curated by professional Photoshop designers)~\cite{ciplab2020deepfake}. The resulting dataset contains around 80,000 instances for \textit{fake} class and 18,000 for \textit{real} one. For each instance, facial regions were automatically detected and cropped from video frames, following protocols introduced in~\cite{pintelas2025quantization}. This task reflects real-world concerns in digital media forensics and misinformation detection.

\subsubsection{Plant Disease recognition}
A curated plant disease dataset merged from 14 publicly available sources \cite{dobrovsky2022plant} was utilized, combining both laboratory and field images to ensure diversity in acquisition conditions. The final version includes approximately 57,000 samples for \textit{diseased} and 20,000 for \textit{healthy} class. This task reflects practical agricultural challenges in crop health monitoring and early disease detection.

\subsubsection{Pneumonia detection}
The Chest X-ray dataset~\cite{kermany2018identifying} was utilized, which contains 5,856 validated chest radiographs from pediatric patients aged 1–5 years. Images were initially categorized into three labels: \textit{normal}, \textit{bacterial pneumonia}, and \textit{viral pneumonia}. For our benchmark, we formulated a binary classification task by grouping all pneumonia types (4,200 cases) as \textit{pneumonia} and 1,500 \textit{normal} cases. This task reflects a high-impact real-world application in pediatric respiratory disease diagnosis using medical imaging.

\section{VO-aware Evaluation Protocol: Ratio-Constrained JSD-guided Clustering}\label{vo-aware-eval}

The proposed evaluation protocol simulates a limited-data regime, where the training set comprises only a small portion of the total data (e.g., 30\%), and exhibits a strong distributional shift from the remaining 70\% testing set. This setting reflects real-world and competition-style environments where such imbalances and shifts frequently occur \cite{zhao2024breaking,xu2024cbdmoe}.

To simulate this, we introduce a ratio-constrained, JSD-guided partitioning method, which generates maximally divergent train/test splits. Rather than performing random sampling, our method explicitly optimizes the separation between distributions of the two subsets, quantified by the Jensen-Shannon Divergence, while strictly preserving a user-defined class-wise ratio (e.g., 30-70). Importantly, to prevent inter-class drift and avoid biasing the dataset class distribution, the procedure is applied independently for each class.

Let $\mathcal{X} = \{(x_i, y_i)\}_{i=1}^N$ be the dataset with softmax vectors $x_i \in \mathbb{R}^C$ and class labels $y_i \in \mathcal{Y}$. For each class $c \in \mathcal{Y}$, define $\mathcal{X}_c = \{x_i \mid y_i = c\}$ with $N_c = |\mathcal{X}_c|$.
We seek a partition of $\mathcal{X}_c$ into disjoint subsets $\mathcal{X}_{T}^c$ and $\mathcal{X}_{E}^c$ of fixed ratio $r$ such that the Jensen-Shannon Divergence between their distributions is maximized:
\[
\mathcal{X}_{T}^c, \mathcal{X}_{E}^c = \arg\max_{\substack{|\mathcal{X}_{T}^c| = r N_c \\ |\mathcal{X}_{E}^c| = (1 - r) N_c}} \text{JSD}\left(\mu_{T}^c \parallel \mu_{E}^c\right),
\]
where \(
\mu_{T}^c = \frac{1}{|\mathcal{X}_{T}^c|} \sum_{x_i \in \mathcal{X}_{T}^c} x_i \quad
\text{and} 
\quad
\mu_{E}^c = \frac{1}{|\mathcal{X}_{E}^c|} \sum_{x_i \in \mathcal{X}_{E}^c} x_i 
\)
are the mean softmax vectors per split.
The final global partition is given by:
\(
\mathcal{X}_{T} = \bigcup_{c \in \mathcal{Y}} \mathcal{X}_{T}^c, \quad \text{ and }\quad \mathcal{X}_{E} = \bigcup_{c \in \mathcal{Y}} \mathcal{X}_{E}^c.
\)

\noindent
This optimization problem defines the underlying objective of our partitioning scheme. To approximate this solution in practice, we employ an iterative refinement strategy that incrementally reassigns data points based on their relative divergence to the evolving training and test centroids. The complete procedure is detailed in Algorithm \ref{alg:jsd_cluster}.

\begin{algorithm}[!ht]
\small{
\caption{Ratio-constrained JSD-guided clustering}
\label{alg:jsd_cluster}
\textbf{Input:} Softmax vectors $\mathcal{X} = \{(x_i, y_i)\}_{i=1}^N$; target ratio $r$; maximum iterations $T$; optional overlap $\zeta \in [0, 0.5]$ (default $=0$) \\
\textbf{Output:} Training set indices $\mathcal{I}_{T}$; Testing set indices $\mathcal{I}_{E}$

\begin{algorithmic}[1]
\STATE Initialize $\mathcal{I}_{T} \leftarrow \emptyset$, $\mathcal{I}_{E} \leftarrow \emptyset$
\FORALL{class $c \in \mathcal{Y}$}
    \STATE Let $\mathcal{X}_c = \{x_i \mid y_i = c\}$, total $N_c = |\mathcal{X}_c|$
    \STATE Randomly select $r \cdot N_c$ samples as initial $\mathcal{I}_{T}^c$ and the rest as $\mathcal{I}_{E}^c$
    \FOR{iteration = $1$ to $T$}
        \STATE $\mu_{T}^c \leftarrow \text{mean}(\mathcal{X}_{\mathcal{I}_{T}^c})$, $\mu_{E}^c \leftarrow \text{mean}(\mathcal{X}_{\mathcal{I}_{E}^c})$
        \FORALL{$x_i \in \mathcal{X}_c$}
            \STATE $d_{T}(x_i) \leftarrow \text{JSD}(x_i, \mu_{T}^c)$; \quad $d_{E}(x_i) \leftarrow \text{JSD}(x_i, \mu_{E}^c)$; \quad $\Delta_i \leftarrow d_{E}(x_i) - d_{T}(x_i)$
        \ENDFOR
        \STATE Sort $x_i$ by $\Delta_i$ and reassign top $r \cdot N_c$ to $\mathcal{I}_{T}^c$, rest to $\mathcal{I}_{E}^c$
        \STATE \textbf{if} global JSD gain $< \epsilon$ \textbf{break}
    \ENDFOR
\IF{$\zeta > 0$}
    %\STATE \textit{// optional user-steered overlap}
    \STATE $s_c \leftarrow \left\lfloor \zeta \cdot |\mathcal{I}_{T}^c| \right\rfloor$; 
           sample $S_T^c \subset \mathcal{I}_{T}^c$, $|S_T^c|=s_c$; 
           sample $S_E^c \subset \mathcal{I}_{E}^c$, $|S_E^c|=s_c$
    \STATE $\mathcal{I}_{T}^c \leftarrow (\mathcal{I}_{T}^c \setminus S_T^c)\, \cup\, S_E^c$;
           $\mathcal{I}_{E}^c \leftarrow (\mathcal{I}_{E}^c \setminus S_E^c)\, \cup\, S_T^c$
\ENDIF
\STATE Append $\mathcal{I}_{T}^c$ to $\mathcal{I}_{T}$, and $\mathcal{I}_{E}^c$ to $\mathcal{I}_{E}$
\ENDFOR
\end{algorithmic}
}
\end{algorithm}

The input to the algorithm consists of softmax vector representations of all data instances. These representations are extracted via a fixed, unbiased encoder \( Z \), such as a frozen ImageNet-pretrained network, to ensure that no task-specific adaptation influences the embeddings. In addition, the input parameter ``target ratio $r$'' controls the fraction of samples assigned to the training subset per class (e.g., $r = 0.3$ corresponds to 30\% training and 70\% testing sets ratios). This parameter reflects the severity of the limited data condition. The lower the training ratio, the more severe the data scarcity becomes, thereby amplifying the risk of validation overfitting.

\vspace{-\parskip}
The algorithm proceeds by iterating over each class $c \in \mathcal{Y}$ individually. In Step 3, for each class, the subset $\mathcal{X}_c$ of all instances belonging to that class is extracted. 
% , ensuring that class-specific distributions are handled independently. This intra-class handling is essential for preserving the dataset's inherent class distribution and avoiding biased splits that might arise from inter-class separation. 
In Step 4, an initial random split is created within $\mathcal{X}_c$, assigning exactly $r \cdot N_c$ samples to the training subset $\mathcal{I}_{T}^c$ and the remaining $(1-r) \cdot N_c$ to the test subset $\mathcal{I}_{E}^c$, where $N_c$ is the number of instances in subset $\mathcal{X}_c$.

\vspace{-\parskip}
The main body of the algorithm (Steps 5-12) is an iterative refinement loop, executed for a fixed number of iterations $T$ or until convergence. In each iteration, the algorithm in Step 6, first computes the mean softmax vectors of the current training and test subsets, denoted by $\mu_{T}^c$ and $\mu_{E}^c$, respectively. These centroids serve as reference points for evaluating how well each instance aligns distributionally with either subset. Next, in Steps 7-9, for every instance $x_i \in \mathcal{X}_c$, two JSD values are computed: the divergence $d_{T}(x_i)$ between $x_i$ and the training centroid $\mu_{T}^c$, and the divergence $d_{E}(x_i)$ between $x_i$ and the test centroid $\mu_{E}^c$. The difference $\Delta_i = d_{E}(x_i) - d_{T}(x_i)$ is used to quantify the relative affinity of $x_i$ toward the training or testing distribution. A positive $\Delta_i$ implies that $x_i$ is closer to the training centroid than to the test centroid, while a negative value suggests the opposite.

\vspace{-\parskip}
Once all $\Delta_i$ values are computed, the algorithm in Step 10, ranks all instances in $\mathcal{X}_c$ based on their $\Delta_i$ scores. The top $r \cdot N_c$ instances with the highest $\Delta_i$ are reassigned to the training subset $\mathcal{I}_{T}^c$, and the remainder to the test subset $\mathcal{I}_{E}^c$. This reassignment attempts to maximize the separation between the two subsets by moving instances that are more aligned with the training centroid into the training set, and vice versa.

\vspace{-\parskip}
After each iteration, the algorithm in Step 11, checks for convergence by evaluating the change in the global JSD between the updated training and test centroids. This serves as a global divergence score indicating how different the two distributions have become. Importantly, this global JSD must be distinguished from the per-instance JSDs used during reassignment. While the per-instance values guide the local partitioning decisions, the global JSD quantifies the overall quality of the split. If the increase in global JSD between iterations falls below a small threshold $\epsilon$, the process is considered converged, and the loop terminates early.

\vspace{-\parskip}
After the main ratio-constrained phase, which typically yields the maximum achievable JSD at the fixed per-class train/test ratio, we add a single-parameter refinement that allows users to tune the desired shift level (Steps 13-16). Given an overlap ratio $\zeta\in[0,0.5]$, we random sample $s_c=\lfloor \zeta\cdot|\mathcal{I}_T^c|\rfloor$ samples from the minority split and $s_c$ samples from the majority split for each class and swap them. This preserves both the class balance and the train/test ratio while monotonically reducing the global JSD as $\zeta$ increases. Users may vary $\zeta$ (e.g., $0.1,0.2,0.3$) to control the shift strength. For instance, with $|\mathcal{I}_T^c|=3000$, $|\mathcal{I}_E^c|=7000$, and $\zeta=0.2$, we swap $600$ randomly sampled items in each direction. As an illustrative example for Skin-Cancer dataset, with a target ratio $r=0.1$, this refinement decreases the total JSD from $0.445$ (at $\zeta=0$) to $0.316$, $0.229$, $0.163$, $0.110$, and $0.069$ at $\zeta\in{0.1,0.2,0.3,0.4,0.5}$, respectively.

\begin{figure*}[!ht]
\centering
\includegraphics[width=0.90\linewidth]{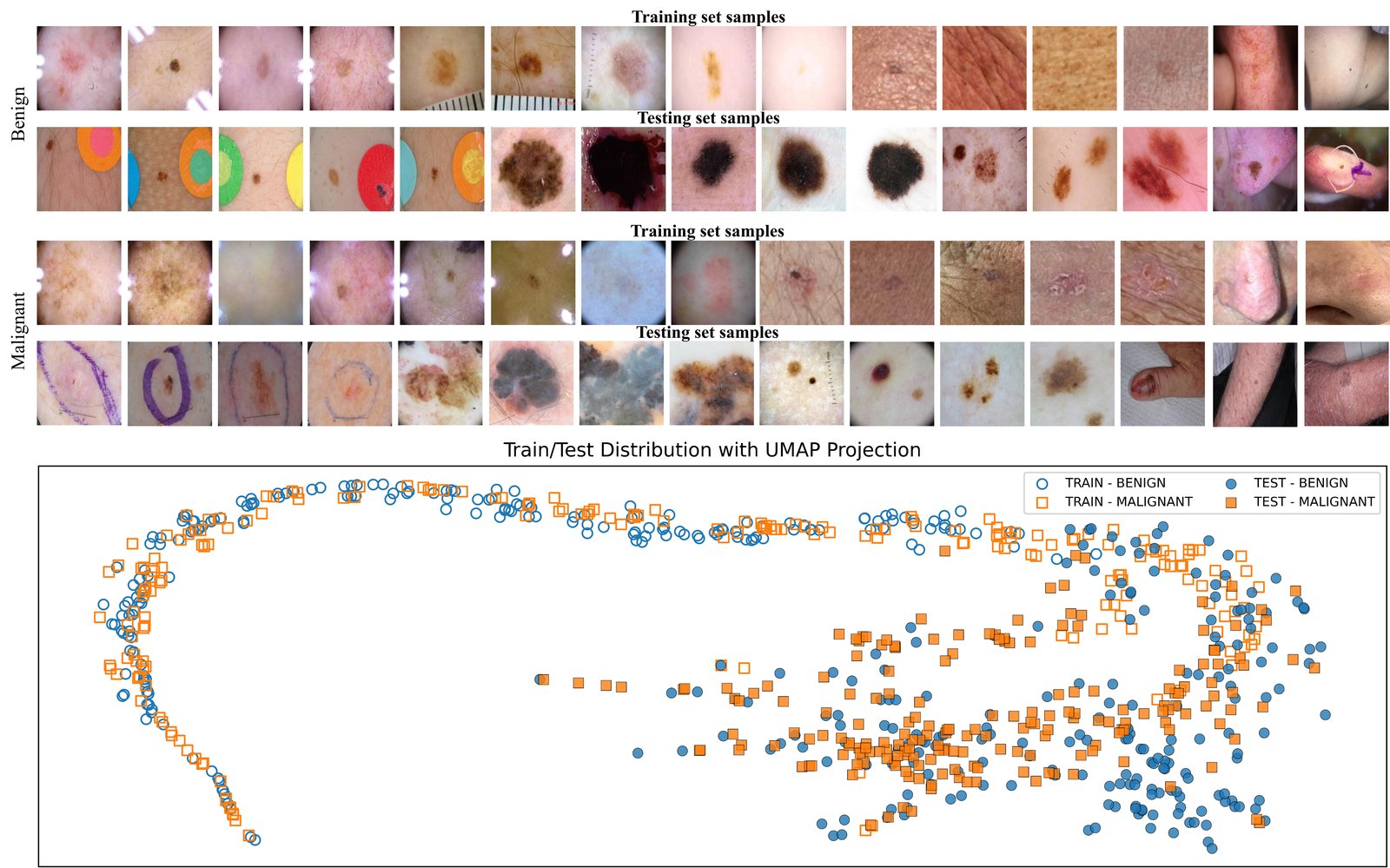}
\caption{Distribution-Shifted Train/Test partitioning based on the proposed JSD-guided clustering algorithm applied on Skin Cancer dataset. The goal of the method is to partition a dataset into training and test subsets such that the JSD between their distributions is maximized. This process generates distributionally shifted splits that simulate realistic and challenging learning scenarios. Such splits are useful for reliably benchmarking a model’s robustness to validation overfitting. Visual inspection reveals distinct data characteristics across the two subsets, emphasizing the diversity of the resulting partitions.}
\label{jsd_partition_skin}
\end{figure*}

\vspace{-\parskip}
Finally, the per-class training and test indices are aggregated into global index sets (Step 17). The algorithm returns these indices, which can then be used to construct the actual data splits.
It is worth emphasizing that this algorithm converges extremely fast in practice, typically requiring only 2 to 3 iterations per class to converge, while produces significantly divergent splits that are highly suitable for VO-aware generalization stress testing. 
% For example, in the Skin Cancer case study dataset, a random 30-70 train-test split yielded a Jensen-Shannon Divergence of only $0.0001$, reflecting nearly identical training and testing distributions. In contrast, the proposed JSD-guided clustering method raised this divergence for the same dataset to $0.36$, effectively simulating a challenging distributional-shifted split. 
Figure~\ref{jsd_partition_skin} visually illustrates the resulting partitioning, where training and test subsets exhibit clearly distinct intra-class patterns, demonstrating the algorithm’s effectiveness in partitioning data into distributionally divergent subsets. 
The full implementation of this algorithm is provided at the following link\footnote{\href{https://github.com/EmmanuelPintelas/-Shifted-Dataset-Creator}{Ratio-constrained JSD-guided clustering}}.

This ability to enforce measurable distributional shift makes the proposed method especially useful for crafting stress-test benchmarks for evaluating generalization and VO behavior.

\section{Experimental Analysis}

This section presents the experimental setup and evaluation results of the proposed GeNeX framework across four challenging real-world binary classification tasks: Skin Cancer detection, DeepFake detection, Plant Disease recognition, and Pneumonia classification. We benchmarked GeNeX against a diverse set of baseline ensemble strategies, under the proposed VO-aware evaluation protocol that deliberately enforces distributional shift to simulate realistic generalization challenges.

\subsection{SoA Baselines Comparison Study}

Table \ref{framework_summary} provides a structured summary of all baseline methods and ensemble configurations involved in the experimental comparison. Each framework is characterized by its model generation strategy, the presence or absence of validation monitoring, the total number of generated models ($N$), the ensemble selection criteria, and the number of final models used for prediction ($K$). We compared GeNeX against the following approaches:

\vspace{-\parskip}
\textit{Single} model selection \cite{sarmah2024ensemble}: Model selection from a pool of $N$ trained networks based on top validation performance. The pool is generated via configuration-driven diversity search, wherein multiple models are trained under varying configurations and the top-performing one on validation is selected. 
\textit{Incremental Ensemble (Inc-Ens)} construction \cite{sarmah2024ensemble}: A sequential ensemble construction strategy that progressively builds an ensemble by selecting $K$ models from a pool of $N$ diverse trained models, with each addition guided by improvements in validation performance.
\textit{Genetic-PneuNet} \cite{kaya2024novel}: A genetic algorithm (GA) based ensemble model originally designed for pneumonia classification. Models are evolved via crossover and mutation, and the ensemble is formed by selecting the best performing offspring based on validation.  % K = 3
\textit{Genetic-SeqSel} \cite{xu2024genetic}: A sequential instance selection framework using genetic algorithms to iteratively build robust ensembles. Though focused on instance selection, the strategy emphasizes validation-guided evolution of ensemble members.  % K = 10
\textit{Genetic-BUSNet} \cite{dar2024deep}: Combines deep learning and genetic algorithms for feature selection and ensemble classification in breast ultrasound image analysis. Emphasizes diversity in the model space with validation-driven selection. % K = 4
\textit{CBDMoE} \cite{xu2024cbdmoe}: A Consistent-But-Diverse Learning (CBDL)-based MoE framework for domain generalization, where experts are trained with consistency and diversity losses, and validation-guided fusion enhances generalizability. 
\textit{PRL} \cite{zhao2024breaking}: A Preference-Vector-driven (PV) hypernetwork framework that generates diverse experts for long-tailed learning, enabling controllable head-tail trade-offs and robust adaptation to testing distribution shifts. 
\textit{Heterogeneity-based eXperts (HeX)} \cite{pintelas2025mobilenet,pintelas2025textnex}: Clusters models by prediction behavior and selects one expert per cluster using different criteria: \textit{HeX-TV} chooses the top-validation performer within each cluster, while \textit{HeX-C} selects the model closest to the cluster centroid in prediction space. 
\textit{RSEN} \cite{xu2022robust}: A robust self-ensembling network that maintains a teacher as the exponential moving average of a supervised student and regularizes training with a consistency loss between teacher and student under stochastic augmentations. The moving-average ensemble network is used as the final predictor; adapted here to 2D convs. \textit{HWA} \cite{gu2023hierarchical}: A hierarchical weight-averaging scheme that periodically averages multiple models trained in parallel (online) and then applies a sliding-window average over saved outer checkpoints (offline), yielding a single robust predictor.

\vspace{-\parskip}
GeNeX differs from the above in that it utilizes validation-independent genetic-based model generation with cluster-wise prototype fusion, aiming to minimize validation reliance both during model generation and ensemble construction.

%\subsubsection{Experimental Settings}
\vspace{-\parskip}
\textit{1) Experimental Settings:}
All base models across methods were trained using the Adam optimizer (initial learning rate ${10}^{-4}$), with reduce-on-plateau scheduling, early stopping (patience = 5), and weight decay (${10}^{-4}$) \cite{song2022learning,borisov2022deep}. For a fair comparison across all methods, we utilized the MobileNetV3-Small pretrained on ImageNet, selected for its strong accuracy-speed trade-off and its effectiveness as a lightweight base learner for ensemble learning~\cite{mobilenetv3}. Input images were uniformly resized to $224 \times 224$ following standard CNN preprocessing practices \cite{zhao2025mixir}. The evaluation metric across all experiments was the Geometric Mean (GM), computed as $\sqrt{\text{sensitivity} \times \text{specificity}}$. GM offers a balanced assessment of classification performance across classes, especially under imbalanced settings \cite{naidu2023review}. All experiments were implemented in PyTorch and executed on NVIDIA RTX 3090 GPUs.

\vspace{-\parskip}
To quantify validation overfitting, we computed the VO gap for each model as the difference between validation and test performance, using the same GM metric. For fairness, all methods used a consistent validation split (30\% of the training pool) and were evaluated against the fixed Shared Test Set. 
% This consistency ensures that observed VO effects stem from methodological differences, not data sampling artifacts.
Table \ref{vo_results} reports the test performance (TP) and validation overfitting (VO) gap for each framework across the four benchmark tasks. TP values correspond to GM performance, and VO gap is the GM difference $(\text{val} - \text{test})$, both scaled by $10^{3}$. The evaluation was performed under two distinct partitioning strategies: (i) Random Split (RS), where the training-test split is created via classic stratified random sampling, resulting in near-zero JSD and thus low VO risk; and (ii) Guided Split (GS) with maximized distributional shift (e.g., $\text{JSD} \in [0.1, 0.6]$ depending on dataset) via the proposed JSD-guided partitioning algorithm (described in Section~\ref{vo-aware-eval}).

At this point, it is essential to clarify that our JSD splitter is a benchmarking utility designed to simulate shifted test environments for evaluating VO-resilience across methods. Using a ratio-constrained partitioning procedure, we construct a \emph{train/validation/(shifted) test} split from the same labeled corpus, preserving fixed per-class ratios while maximizing the distributional separation between train/validation and (shifted) test. The resulting ``test'' set is a proxy used solely for VO-aware evaluation; not a production holdout from the deployment distribution.

\vspace{-\parskip}
The GeNeX framework is evaluated with two GenE population sizes, $N=50$ and $N=500$, where $N$ denotes the number of models generated by the GenE module, to assess robustness under small and large candidate populations.

\begin{table*}[!ht]
\centering
\setlength{\tabcolsep}{5pt}
\renewcommand{\arraystretch}{1}	
\caption{Summary classification of all utilized frameworks in the experimental comparison.}\label{framework_summary}
\small{
\begin{tabular}{c|c|cc|cc}
\hline
\multirow{3}{*}{Frameworks} &
  \multirow{3}{*}{Description} &
  \multicolumn{2}{c|}{Generation} &
  \multicolumn{2}{c}{Selection} \\ \cline{3-6} 
 &
   &
  \multicolumn{1}{c|}{\multirow{2}{*}{\begin{tabular}[c]{@{}c@{}}Validation\\ Monitoring\end{tabular}}} &
  \multirow{2}{*}{N} &
  \multicolumn{1}{c|}{\multirow{2}{*}{Criterion}} &
  \multirow{2}{*}{K} \\
 &
   &
  \multicolumn{1}{c|}{} &
   &
  \multicolumn{1}{c|}{} &
   \\ \hline
Single (1-ep) &
  Single model just trained for 1 epoch. &
  \multicolumn{1}{c|}{\multirow{2}{*}{\xmark}} &
  1 &
  \multicolumn{1}{c|}{\multirow{2}{*}{-}} &
  1 \\ \cline{1-2} \cline{4-4} \cline{6-6} 
Aver-Ens (1-ep) &
  Averaging ensemble of random initialized models trained for 1 epochs. &
  \multicolumn{1}{c|}{} &
  3 &
  \multicolumn{1}{c|}{} &
  3 \\ \hline
\multirow{2}{*}{Single ($N_{50}$)} &
  \multirow{4}{*}{\begin{tabular}[c]{@{}c@{}}(1) A pool of $N$ models is trained under diverse configurations\\ using a standard train-validation development pipeline.\\ (2) Validation performance is monitored to guide model development.\\ (3) The top-validation model is selected as final single predictor\end{tabular}} &
  \multicolumn{1}{c|}{\multirow{16}{*}{\cmark}} &
  \multirow{2}{*}{50} &
  \multicolumn{1}{c|}{\multirow{10}{*}{Top-Val-based}} &
  \multirow{4}{*}{1} \\
 &
   &
  \multicolumn{1}{c|}{} &
   &
  \multicolumn{1}{c|}{} &
   \\
\multirow{2}{*}{Single ($N_{500}$)} &
   &
  \multicolumn{1}{c|}{} &
  \multirow{2}{*}{500} &
  \multicolumn{1}{c|}{} &
   \\
 &
   &
  \multicolumn{1}{c|}{} &
   &
  \multicolumn{1}{c|}{} &
   \\ \cline{1-2} \cline{4-4} \cline{6-6} 
Inc-Ens ($N_{50}, K_{3}$) &
  \multirow{3}{*}{\begin{tabular}[c]{@{}c@{}}Ensemble approach that generates $N$ models as in (1) and (2).\\ Incrementally selects $K$ total models from the generated pool,\\ each chosen to progressively enhance validation performance.\end{tabular}} &
  \multicolumn{1}{c|}{} &
  50 &
  \multicolumn{1}{c|}{} &
  \multirow{2}{*}{3} \\
Inc-Ens ($N_{500}, K_{3}$) &
   &
  \multicolumn{1}{c|}{} &
  500 &
  \multicolumn{1}{c|}{} &
   \\
Inc-Ens ($N_{500}, K_{10}$) &
   &
  \multicolumn{1}{c|}{} &
  500 &
  \multicolumn{1}{c|}{} &
  10 \\ \cline{1-2} \cline{4-4} \cline{6-6} 
Genetic-PneuNet &
  Evolves models via GA and selects best offsprings by validation. &
  \multicolumn{1}{c|}{} &
  \multirow{5}{*}{50} &
  \multicolumn{1}{c|}{} &
  3 \\ \cline{1-2} \cline{6-6} 
Genetic-SeqSel &
  Builds ensembles by GA-driven, validation-based model selection &
  \multicolumn{1}{c|}{} &
   &
  \multicolumn{1}{c|}{} &
  10 \\ \cline{1-2} \cline{6-6} 
Genetic-BUSNet &
  Combines GA feature/model selection with validation-based ensemble. &
  \multicolumn{1}{c|}{} &
   &
  \multicolumn{1}{c|}{} &
  4 \\ \cline{1-2} \cline{5-6} 
CBDMoE &
  Trains experts with consistency-diversity loss for domain generalization. &
  \multicolumn{1}{c|}{} &
   &
  \multicolumn{1}{c|}{CBDL-based} &
  \multirow{6}{*}{Auto} \\ \cline{1-2} \cline{5-5}
PRL &
  Uses hypernetworks to generate experts and adapt via preference vectors. &
  \multicolumn{1}{c|}{} &
   &
  \multicolumn{1}{c|}{PV-based} &
   \\ \cline{1-2} \cline{4-5}
HeX-TV ($N_{50}$) &
  \multirow{2}{*}{\begin{tabular}[c]{@{}c@{}}Performs behavioral clustering on the $N$ generated models (1), (2).\\ One expert per cluster is selected based on validation performance.\end{tabular}} &
  \multicolumn{1}{c|}{} &
  50 &
  \multicolumn{1}{c|}{\multirow{2}{*}{Top-Val-based}} &
   \\
HeX-TV ($N_{500}$) &
   &
  \multicolumn{1}{c|}{} &
  500 &
  \multicolumn{1}{c|}{} &
   \\ \cline{1-2} \cline{4-5}
HeX-C ($N_{50}$) &
  \multirow{2}{*}{\begin{tabular}[c]{@{}c@{}}Selects the model closest to the cluster centroid in prediction space,\\ promoting representativeness over raw performance within each cluster.\end{tabular}} &
  \multicolumn{1}{c|}{} &
  50 &
  \multicolumn{1}{c|}{\multirow{2}{*}{Centroid-based}} &
   \\
HeX-C ($N_{500}$) &
   &
  \multicolumn{1}{c|}{} &
  500 &
  \multicolumn{1}{c|}{} &
   \\ \cline{1-6}
RSEN &
  Moving-average teacher with consistency-filtered self-ensembling. &
  \multicolumn{1}{c|}{\multirow{4}{*}{\xmark}} &
  Auto &
  \multicolumn{1}{c|}{-} &
  \multicolumn{1}{c}{\multirow{2}{*}{1}} \\
HWA &
  Online averaging of parallel models with offline slide-window averaging. &
  \multicolumn{1}{c|}{} &
  2--4 &
  \multicolumn{1}{c|}{Averaging} &
   \\ \cline{1-2} \cline{4-6}
GeNeX ($N_{50}$) &
  \multirow{2}{*}{\begin{tabular}[c]{@{}c@{}}Fuses genetically evolved expert models per cluster \\ into robust prototypes.\end{tabular}} &
  \multicolumn{1}{c|}{} &
  50 &
  \multicolumn{1}{c|}{\multirow{2}{*}{Prototype-based}} & \multicolumn{1}{c}{\multirow{2}{*}{3--5}}
   \\
GeNeX ($N_{500}$) &
   &
  \multicolumn{1}{c|}{} &
  500 &
  \multicolumn{1}{c|}{} &
   \\ \hline
\end{tabular}%
}
\end{table*}

\begin{table*}[!ht]
\centering
\setlength{\tabcolsep}{5pt}
\renewcommand{\arraystretch}{1}	
\caption{Performance results under the proposed VO-aware evaluation protocol: Ratio-constrained JSD-guided clustering algorithm was applied to generate highly divergent training-testing splits (target ratio $r=0.3$).}\label{vo_results}
\small{
\begin{tabular}{@{}c|cccc|cccc|cccc|cccc@{}}
\toprule
\multirow{3}{*}{Frameworks} &
  \multicolumn{4}{c|}{Skin Cancer} &
  \multicolumn{4}{c|}{DeepFake} &
  \multicolumn{4}{c|}{Plant Disease} &
  \multicolumn{4}{c}{Pneumonia} \\ \cmidrule(l){2-17} 
 &
  \multicolumn{2}{c|}{\begin{tabular}[c]{@{}c@{}}RS\\ JSD = 1e-4\end{tabular}} &
  \multicolumn{2}{c|}{\begin{tabular}[c]{@{}c@{}}GS\\ JSD = 0.360\end{tabular}} &
  \multicolumn{2}{c|}{\begin{tabular}[c]{@{}c@{}}RS\\ JSD = 1e-4\end{tabular}} &
  \multicolumn{2}{c|}{\begin{tabular}[c]{@{}c@{}}GS\\ JSD = 0.545\end{tabular}} &
  \multicolumn{2}{c|}{\begin{tabular}[c]{@{}c@{}}RS\\ JSD = 1e-4\end{tabular}} &
  \multicolumn{2}{c|}{\begin{tabular}[c]{@{}c@{}}GS\\ JSD = 0.434\end{tabular}} &
  \multicolumn{2}{c|}{\begin{tabular}[c]{@{}c@{}}RS\\ JSD = 1e-4\end{tabular}} &
  \multicolumn{2}{c}{\begin{tabular}[c]{@{}c@{}}GS\\ JSD = 0.142\end{tabular}} \\ \cmidrule(l){2-17} 
 &
  TP &
  \multicolumn{1}{c|}{VO} &
  TP &
  VO &
  TP &
  \multicolumn{1}{c|}{VO} &
  TP &
  VO &
  TP &
  \multicolumn{1}{c|}{VO} &
  TP &
  VO &
  TP &
  \multicolumn{1}{c|}{VO} &
  TP &
  VO \\ \midrule
Single (1-ep) &
  696 &
  \multicolumn{1}{c|}{\textless{}1} &
  672 &
  54 &
  658 &
  \multicolumn{1}{c|}{\textless{}1} &
  585 &
  19 &
  769 &
  \multicolumn{1}{c|}{\textless{}1} &
  597 &
  \textbf{171} &
  526 &
  \multicolumn{1}{c|}{\textless{}1} &
  533 &
  120 \\
Aver-Ens (1-ep) &
  702 &
  \multicolumn{1}{c|}{\textless{}1} &
  693 &
  55 &
  697 &
  \multicolumn{1}{c|}{\textless{}1} &
  605 &
  16 &
  847 &
  \multicolumn{1}{c|}{\textless{}1} &
  600 &
  175 &
  541 &
  \multicolumn{1}{c|}{\textless{}1} &
  645 &
  113 \\ \midrule
Single ($N_{50}$) &
  791 &
  \multicolumn{1}{c|}{2} &
  607 &
  214 &
  706 &
  \multicolumn{1}{c|}{3} &
  507 &
  234 &
  943 &
  \multicolumn{1}{c|}{\textless{}1} &
  533 &
  380 &
  914 &
  \multicolumn{1}{c|}{\textless{}1} &
  812 &
  152 \\
Single ($N_{500}$) &
  802 &
  \multicolumn{1}{c|}{8} &
  577 &
  248 &
  715 &
  \multicolumn{1}{c|}{9} &
  445 &
  298 &
  946 &
  \multicolumn{1}{c|}{2} &
  540 &
  395 &
  929 &
  \multicolumn{1}{c|}{\textless{}1} &
  798 &
  170 \\
Inc-Ens ($N_{50}, k_{3}$) &
  813 &
  \multicolumn{1}{c|}{13} &
  623 &
  236 &
  741 &
  \multicolumn{1}{c|}{11} &
  429 &
  334 &
  949 &
  \multicolumn{1}{c|}{5} &
  504 &
  434 &
  932 &
  \multicolumn{1}{c|}{\textless{}1} &
  806 &
  164 \\
Inc-Ens ($N_{500}, k_{3}$) &
  \textbf{825} &
  \multicolumn{1}{c|}{15} &
  597 &
  267 &
  752 &
  \multicolumn{1}{c|}{17} &
  401 &
  397 &
  952 &
  \multicolumn{1}{c|}{8} &
  498 &
  443 &
  946 &
  \multicolumn{1}{c|}{\textless{}1} &
  793 &
  196 \\
Inc-Ens ($N_{500}, k_{10}$) &
  810 &
  \multicolumn{1}{c|}{38} &
  584 &
  295 &
  760 &
  \multicolumn{1}{c|}{36} &
  404 &
  446 &
  942 &
  \multicolumn{1}{c|}{27} &
  405 &
  573 &
  \textbf{963} &
  \multicolumn{1}{c|}{3} &
  711 &
  288 \\ \midrule
Genetic-PneuNet &
  814 &
  \multicolumn{1}{c|}{7} &
  659 &
  142 &
  771 &
  \multicolumn{1}{c|}{8} &
  556 &
  157 &
  950 &
  \multicolumn{1}{c|}{2} &
  574 &
  346 &
  \textbf{962} &
  \multicolumn{1}{c|}{\textless{}1} &
  826 &
  138 \\
Genetic-SeqSel &
  821 &
  \multicolumn{1}{c|}{12} &
  646 &
  162 &
  768 &
  \multicolumn{1}{c|}{9} &
  571 &
  175 &
  953 &
  \multicolumn{1}{c|}{3} &
  566 &
  379 &
  952 &
  \multicolumn{1}{c|}{\textless{}1} &
  823 &
  139 \\
Genetic-BUSNet &
  819 &
  \multicolumn{1}{c|}{8} &
  683 &
  101 &
  777 &
  \multicolumn{1}{c|}{9} &
  591 &
  181 &
  962 &
  \multicolumn{1}{c|}{1} &
  558 &
  383 &
  955 &
  \multicolumn{1}{c|}{\textless{}1} &
  819 &
  156 \\
CBDMoE &
  815 &
  \multicolumn{1}{c|}{3} &
  696 &
  99 &
  772 &
  \multicolumn{1}{c|}{4} &
  567 &
  80 &
  954 &
  \multicolumn{1}{c|}{\textless{}1} &
  641 &
  307 &
  958 &
  \multicolumn{1}{c|}{\textless{}1} &
  830 &
  148 \\
PRL &
  809 &
  \multicolumn{1}{c|}{2} &
  701 &
  87 &
  \textbf{793} &
  \multicolumn{1}{c|}{1} &
  617 &
  52 &
  \textbf{969} &
  \multicolumn{1}{c|}{\textless{}1} &
  667 &
  203 &
  950 &
  \multicolumn{1}{c|}{\textless{}1} &
  838 &
  112 \\
HeX-TV ($N_{50}$) &
  \textbf{831} &
  \multicolumn{1}{c|}{9} &
  675 &
  192 &
  \textbf{790} &
  \multicolumn{1}{c|}{5} &
  615 &
  155 &
  967 &
  \multicolumn{1}{c|}{3} &
  583 &
  367 &
  959 &
  \multicolumn{1}{c|}{\textless{}1} &
  817 &
  170 \\
HeX-TV ($N_{500}$) &
  822 &
  \multicolumn{1}{c|}{20} &
  634 &
  242 &
  779 &
  \multicolumn{1}{c|}{21} &
  572 &
  244 &
  \textbf{972} &
  \multicolumn{1}{c|}{6} &
  552 &
  411 &
  961 &
  \multicolumn{1}{c|}{1} &
  812 &
  182 \\
HeX-C ($N_{50}$) &
  810 &
  \multicolumn{1}{c|}{8} &
  698 &
  98 &
  786 &
  \multicolumn{1}{c|}{5} &
  595 &
  45 &
  965 &
  \multicolumn{1}{c|}{\textless{}1} &
  628 &
  284 &
  953 &
  \multicolumn{1}{c|}{\textless{}1} &
  840 &
  \textbf{103} \\
HeX-C ($N_{500}$) &
  817 &
  \multicolumn{1}{c|}{10} &
  660 &
  167 &
  785 &
  \multicolumn{1}{c|}{8} &
  576 &
  123 &
  963 &
  \multicolumn{1}{c|}{2} &
  607 &
  339 &
  954 &
  \multicolumn{1}{c|}{\textless{}1} &
  831 &
  106 \\
RSEN &
  806 &
  \multicolumn{1}{c|}{\textless{}1} &
  \textbf{721} &
  \textbf{39} &
  780 &
  \multicolumn{1}{c|}{\textless{}1} &
  622 &
  18 &
  948 &
  \multicolumn{1}{c|}{\textless{}1} &
  699 &
  175 &
  956 &
  \multicolumn{1}{c|}{\textless{}1} &
  847 &
  102 \\
HWA &
  812 &
  \multicolumn{1}{c|}{\textless{}1} &
  706 &
  45 &
  783 &
  \multicolumn{1}{c|}{\textless{}1} &
  631 &
  \textbf{15} &
  956 &
  \multicolumn{1}{c|}{\textless{}1} &
  \textbf{702} &
  172 &
  943 &
  \multicolumn{1}{c|}{\textless{}1} &
  \textbf{856} &
  \textbf{95} \\ \midrule
GeNeX ($N_{50}$) &
  816 &
  \multicolumn{1}{c|}{\textless{}1} &
  \textbf{718} &
  46 &
  788 &
  \multicolumn{1}{c|}{\textless{}1} &
  \textbf{643} &
  \textbf{13} &
  947 &
  \multicolumn{1}{c|}{\textless{}1} &
  \textbf{705} &
  \textbf{167} &
  947 &
  \multicolumn{1}{c|}{\textless{}1} &
  \textbf{861} &
  \textbf{87} \\
GeNeX ($N_{500}$) &
  823 &
  \multicolumn{1}{c|}{\textless{}1} &
  716 &
  \textbf{41} &
  784 &
  \multicolumn{1}{c|}{\textless{}1} &
  \textbf{639} &
  17 &
  949 &
  \multicolumn{1}{c|}{\textless{}1} &
  698 &
  183 &
  951 &
  \multicolumn{1}{c|}{\textless{}1} &
  852 &
  105 \\ \bottomrule
\end{tabular}%
}
\end{table*}

%\subsubsection{Discussion of Results}
\vspace{-\parskip}
\textit{2) Discussion of Results:}
The experimental results across all four case study tasks confirm the effectiveness of the proposed GeNeX framework in mitigating validation overfitting and enhancing generalization.
In RS settings (JSD $\approx 0$), where the validation and testing distributions are nearly identical, most methods perform comparably well, with minimal VO gaps. This is expected, as the absence of distributional shift reduces the risk of overfitting, making generalization straightforward.
However, in the GS conditions, where JSD is deliberately maximized to simulate realistic distributional shifts, most models that performed well in RS settings now experience a significant drop in performance. Interestingly, the 1-epoch baselines exhibit strong robustness, consistently yielding low VO gaps and competitive test performance, particularly in the most challenging high-JSD tasks (JSD $> 0.3$). This indicates that extremely short training, without validation monitoring, acts as a natural regularizer, limiting overfitting and enhancing robustness to distributional shift.
GeNeX outperforms all baseline frameworks under GS settings in both test performance and VO gap. This highlights its robustness in scenarios where validation-based model selection becomes unreliable. Notably, even under large model generation budgets (e.g., $N = 500$), GeNeX maintains minimal VO gaps, confirming that its validation-independent generation phase (GenE) and prototype-based ensemble construction (ProtoNeX) effectively prevents validation-induced overfitting.
In contrast, Inc-Ens and top-validation HeX variants show increasing VO gaps as $N$ and $K$ scale up. While their performance is strong in RS, these methods collapse under high JSD conditions in the GS setup. This aligns with the theoretical formulation in Section~\ref{vo_definition}, where we showed that validation-driven model search and ensemble expansion magnify VO risk, which is further amplified in high-JSD regimes.

\vspace{-\parskip}
Among the SoA baselines, genetic strategies like Genetic-PneuNet and Genetic-SeqSel offer improved VO resistance but still suffer under GS due to partial reliance on validation during selection. Likewise, CBDMoE and PRL demonstrate strong accuracy but display moderate VO gaps, particularly in the most shifted scenarios, highlighting the limitations of diversity-driven designs without validation control. HWA maintains low VO and stable generalization in both RS and GS. Its validation-independent hierarchical averaging acts as a strong regularizer, mitigating shift-induced variance without relying on validation-driven search. RSEN leverages self-ensembling regularization to temper validation overfitting and narrow VO gaps under shift. It remains competitive across settings, though its gains are generally milder than GeNeX in the most shifted regimes.

\vspace{-\parskip}
Overall, GeNeX provides robust performance and minimal VO gaps across all settings, outperforming all other methods in the high-risk GS conditions. Its resilience to validation-induced overfitting, without compromising TP, establishes it as a reliable framework for real-world deployments where validation signals are noisy, biased, or unrepresentative.

% \subsubsection{Ablations and statistics}
\vspace{-\parskip}
\textit{3) Ablations and statistics:} We conduct three studies to assess statistical significance, component utility, and hyperparameter robustness. We first apply a nonparametric Friedman Aligned-Ranks test \cite{garcia2010advanced} of the null hypothesis $H_0$ (equal test performance across methods); if $H_0$ is rejected (meaning that the performance differences are not due to random variation), we run Finner post-hoc test comparisons at significance level $\alpha = 5\%$ to examine the existence of significant differences. Next, we perform a ProtoNeX leave-one-out ablation over its expert-selection criteria to quantify each criterion’s contribution. Finally, we sweep GenE’s mutation probability $\rho$ and noise scale $\sigma$ to probe sensitivity. Full protocols and tables appear in the Supplement.\footnote{Supplementary Material (\emph{supplementary.pdf}): Secs.~S1–S3, Tabs.~S1–S3.} The statistical test rejects $H_0$ and post-hoc results show GeNeX to consistently surpasses the strongest baselines. In the ProtoNeX ablation, removing any single criterion increases VO and lowers TP, with the largest drop occurring when the robust-to-perturbations criterion is omitted. In GenE, performance is stable for $\rho\!\in\![0.02,0.10]$ and $\sigma\!\in\![5\!\times\!10^{-3},10^{-2}]$; very small values reduce diversity, whereas larger values can destabilize offspring.

\subsection{Inner Study: Genetic process as Overfitting-Resilient path}

\begin{figure*}[!ht]
\centering
\includegraphics[width=0.90\linewidth]{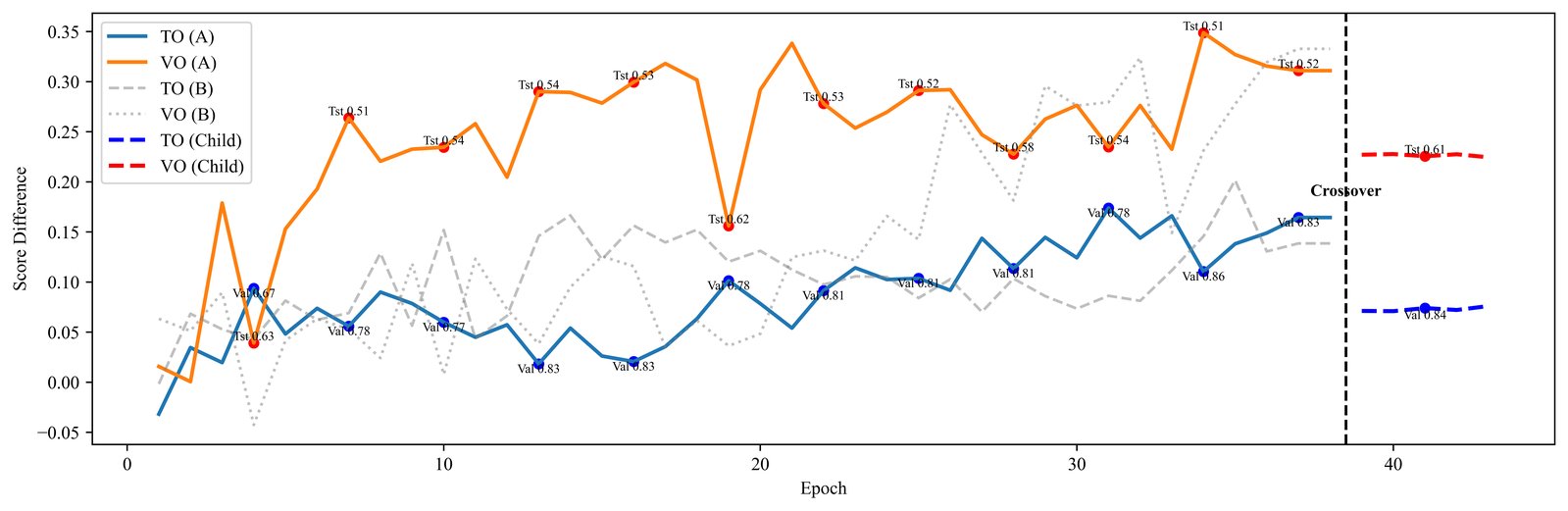}
\caption{Visualization of Train–Val Overfitting (TO) and Val–Test Overfitting (VO) gaps for two parent networks A and B and their genetically evolved child models. The study is conducted on the most challenging VO-pruned dataset, GS-DeepFake, which exhibits the highest distributional shift (JSD = 0.545). We observe that despite minimal TO and high validation performance (epochs 12--18 for A), severe VO gaps can still occur, leading to unreliable selection. After genetic crossover (vertical line), child models demonstrate improved generalization and reduced overfitting.}
\label{fig:inner_view_study}
\end{figure*}

To better understand the internal behavior of GeNeX, we conduct an in-depth study on the GS-DeepFake task, our most challenging VO-pruned benchmark with a pronounced distributional shift (JSD = 0.545). This inner-view analysis focuses on how genetic model generation mitigates both training and validation overfitting. Specifically, we investigate how two overfitted parent models, A and B (Figure~\ref{fig:inner_view_study}), evolve into more generalizable child models through genetic operations.
To simulate the effect of overfitting, we deliberately train parent networks A and B for extended epochs, exposing them to both high Train–Val (TO) and Val–Test (VO) divergence. Then, apply weight-level crossover and mutation to generate child networks, followed by light fine-tuning limited only to the classification head. Our hypothesis is that this genetic crossover process acts as a weight-space regularizer, reinitializing over-optimized trajectories and promoting the exploration of new parameter regions with improved generalization.
While TO can be visible during training and typically correlates with poor generalization, VO is more deceptive. 

We observe that models exhibiting minimal TO and high validation scores (typically considered promising checkpoints) can still fail dramatically at test time. For instance, around in the 12–18 epochs area for A, validation accuracy appears stable and strong, yet large VO gaps exist, highlighting the risk of relying solely on validation performance under distributional shift.
After the crossover point (dashed vertical line), we evaluate the generated offspring/child model. Despite being derived from heavily overfitted parents, the child model consistently show smaller TO and VO gaps and stronger test performance. 

%\vspace{-\parskip}
Overall, this analysis reinforces our claim (Section~\ref{study2}) that genetic operations within GenE serve as a recovery mechanism, which provides an implicit regularization effect mitigating both TO and VO.

\vspace{-0.5ex}
\section{Conclusions}\label{sec:conclusion}

Validation overfitting represents a critical yet often overlooked failure mode in modern deep learning pipelines, especially in ensemble learning settings where model selection and aggregation rely heavily on validation metrics. This paper introduced GeNeX, a framework designed to explicitly address VO by decoupling both model generation and ensemble construction from validation dependence.
GeNeX operates in two distinct phases: the \textit{GenE} module creates a diverse pool of candidate networks through a hybrid supervised–genetic strategy that bypasses validation monitoring entirely; the \textit{ProtoNeX} module then clusters models based on predictive behavior, selects complementary experts within each cluster, and fuses their weights into robust cluster prototypes.
By minimizing reliance on validation queries and enforcing both representational and behavioral diversity, GeNeX achieves strong generalization while systematically reducing VO gaps, even under extreme distribution shifts and large ensemble scales. Theoretical analysis and VO-aware evaluation protocols empirically support its robustness against both model-induced and dataset-induced VO risk.
Additionally, extensive experiments across four real-world classification tasks, Skin Cancer, DeepFake, Plant Disease and Pneumonia, demonstrate that GeNeX consistently outperforms SoA ensemble models in VO-prone scenarios. In addition to its strong performance, GeNeX demonstrates notable robustness under distributional shift, making it suitable for real-world deployments where validation overfitting is an inherent and critical risk.

\vspace{-\parskip}
\textit{Limitations and Future Work:} 
While GeNeX has been validated on challenging real-world case study tasks, the scope of this work remains limited on the image classification domain. However, GeNeX is architecture- and modality-agnostic because it operates on model logits and behavior-driven predictive signatures rather than raw inputs; thus, only the base learner change, while the GenE search and ProtoNeX clustering/prototyping logic remains the same. Future directions include extending the framework to other modalities such as segmentation and text classification. Concretely, for semantic segmentation we would replace the base learner with a dense predictor such as U-Net, keep GenE unchanged as it breeds offspring models via weight fusion of task-trained bases, and apply ProtoNeX to per-pixel or pooled logits. For text classification the base becomes a transformer classifier, GenE again recombines base weights to generate offspring, and ProtoNeX groups models by sequence-level logits. Finally, the proposed VO-aware evaluation protocol could serve as a standardized benchmark for assessing generalization robustness under realistic distributional-shifted environments, encouraging future research in this critical area.

\bibliographystyle{IEEEtran}

%%\bibliography{bibliography}%loading bibliography.bib external file

\begin{thebibliography}{10}
\providecommand{\url}[1]{#1}
\csname url@samestyle\endcsname
\providecommand{\newblock}{\relax}
\providecommand{\bibinfo}[2]{#2}
\providecommand{\BIBentrySTDinterwordspacing}{\spaceskip=0pt\relax}
\providecommand{\BIBentryALTinterwordstretchfactor}{4}
\providecommand{\BIBentryALTinterwordspacing}{\spaceskip=\fontdimen2\font plus
\BIBentryALTinterwordstretchfactor\fontdimen3\font minus
  \fontdimen4\font\relax}
\providecommand{\BIBforeignlanguage}[2]{{%
\expandafter\ifx\csname l@#1\endcsname\relax
\typeout{** WARNING: IEEEtran.bst: No hyphenation pattern has been}%
\typeout{** loaded for the language `#1'. Using the pattern for}%
\typeout{** the default language instead.}%
\else
\language=\csname l@#1\endcsname
\fi
#2}}
\providecommand{\BIBdecl}{\relax}
\BIBdecl

\bibitem{pintelas2025mobilenet}
E.~Pintelas, I.~E. Livieris, V.~Tampakas, and P.~Pintelas, ``{MobileNet-HeX}:
  Heterogeneous ensemble of mobilenet experts for efficient and scalable vision
  model optimization,'' \emph{Big Data and Cognitive Computing}, vol.~9, no.~1,
  p.~2, 2025.

\bibitem{zhao2024breaking}
Z.~Zhao, H.~Wen, Z.~Wang, P.~Wang, F.~Wang, S.~Lai, Q.~Zhang, and Y.~Wang,
  ``Breaking long-tailed learning bottlenecks: {A} controllable paradigm with
  hypernetwork-generated diverse experts,'' \emph{Advances in Neural
  Information Processing Systems}, vol.~37, pp. 7493--7520, 2024.

\bibitem{xu2024cbdmoe}
F.~Xu, D.~Chen, T.~Jia, S.~Deng, and H.~Wang, ``Cbdmoe: Consistent-but-diverse
  mixture of experts for domain generalization,'' \emph{IEEE Transactions on
  Multimedia}, 2024.

\bibitem{sarmah2024ensemble}
U.~Sarmah, P.~Borah, and D.~K. Bhattacharyya, ``Ensemble learning methods: {An}
  empirical study,'' \emph{SN Computer Science}, vol.~5, no.~7, p. 924, 2024.

\bibitem{pintelas2025textnex}
E.~Pintelas, A.~Koursaris, I.~E. Livieris, and V.~Tampakas, ``{TextNeX}: {Text}
  network of experts for robust text classification—case study on
  machine-generated-text detection,'' \emph{Mathematics}, vol.~13, no.~10, p.
  1555, 2025.

\bibitem{xu2024genetic}
C.~Xu and S.~Zhang, ``A genetic algorithm-based sequential instance selection
  framework for ensemble learning,'' \emph{Expert Systems with Applications},
  vol. 236, p. 121269, 2024.

\bibitem{kaya2024novel}
M.~Kaya and Y.~{\c{C}}etin-Kaya, ``A novel ensemble learning framework based on
  a genetic algorithm for the classification of pneumonia,'' \emph{Engineering
  Applications of Artificial Intelligence}, vol. 133, p. 108494, 2024.

\bibitem{dar2024deep}
M.~F. Dar and A.~Ganivada, ``Deep learning and genetic algorithm-based ensemble
  model for feature selection and classification of breast ultrasound images,''
  \emph{Image and Vision Computing}, vol. 146, p. 105018, 2024.

\bibitem{englesson2021generalized}
E.~Englesson and H.~Azizpour, ``Generalized jensen-shannon divergence loss for
  learning with noisy labels,'' \emph{Advances in Neural Information Processing
  Systems}, vol.~34, pp. 30\,284--30\,297, 2021.

\bibitem{gaudreault2024empirical}
J.-G. Gaudreault and P.~Branco, ``Empirical analysis of performance assessment
  for imbalanced classification,'' \emph{Machine Learning}, pp. 1--43, 2024.

\bibitem{aburass2024quantifying}
S.~Aburass and M.~A. Rumman, ``Quantifying overfitting: introducing the
  overfitting index,'' in \emph{2024 International Conference on Electrical,
  Computer and Energy Technologies (ICECET}.\hskip 1em plus 0.5em minus
  0.4em\relax IEEE, 2024, pp. 1--7.

\bibitem{blm2013}
S.~Boucheron, G.~Lugosi, and P.~Massart, \emph{Concentration Inequalities: A
  Nonasymptotic Theory of Independence}.\hskip 1em plus 0.5em minus 0.4em\relax
  Oxford, UK: Oxford University Press, 02 2013.

\bibitem{snapshot_ensemble}
G.~Huang, Y.~Li, G.~Pleiss, Z.~Liu, J.~Hopcroft, and K.~Q. Weinberger,
  ``Snapshot ensembles: {Train} 1, get m for free,'' \emph{International
  Conference on Learning Representations (ICLR)}, 2017.

\bibitem{swa_ensemble}
P.~Izmailov, D.~Podoprikhin, T.~Garipov, D.~Vetrov, and A.~G. Wilson,
  ``Averaging weights leads to wider optima and better generalization,''
  \emph{Conference on Uncertainty in Artificial Intelligence (UAI)}, 2018.

\bibitem{virtanen2020scipy}
P.~Virtanen, R.~Gommers, T.~E. Oliphant \emph{et~al.}, ``Scipy 1.0: Fundamental
  algorithms for scientific computing in python,'' \emph{Nature Methods},
  vol.~17, no.~3, pp. 261--272, 2020.

\bibitem{zhang2021gaussian}
Y.~Zhang, M.~Li, S.~Wang, S.~Dai, L.~Luo, E.~Zhu, H.~Xu, X.~Zhu, C.~Yao, and
  H.~Zhou, ``Gaussian mixture model clustering with incomplete data,''
  \emph{ACM Transactions on Multimedia Computing, Communications, and
  Applications (TOMM)}, vol.~17, no.~1s, pp. 1--14, 2021.

\bibitem{choi2020combining}
J.~Y. Choi and B.~Lee, ``Combining of multiple deep networks via ensemble
  generalization loss, based on mri images, for alzheimer's disease
  classification,'' \emph{IEEE Signal processing letters}, vol.~27, pp.
  206--210, 2020.

\bibitem{ijcai2020p429}
T.~T. Nguyen, N.~Ye, and P.~Bartlett, ``Greedy convex ensemble,'' in
  \emph{Proceedings of the Twenty-Ninth International Joint Conference on
  Artificial Intelligence, {IJCAI-20}}, 7 2020, pp. 3101--3107.

\bibitem{garipov2018loss}
T.~Garipov, P.~Izmailov, D.~Podoprikhin, D.~P. Vetrov, and A.~G. Wilson, ``Loss
  surfaces, mode connectivity, and fast ensembling of dnns,'' \emph{Advances in
  neural information processing systems}, vol.~31, 2018.

\bibitem{wortsman2022model}
M.~Wortsman, G.~Ilharco, S.~Y. Gadre, R.~Roelofs, R.~Gontijo-Lopes, A.~S.
  Morcos, H.~Namkoong, A.~Farhadi, Y.~Carmon, S.~Kornblith \emph{et~al.},
  ``Model soups: averaging weights of multiple fine-tuned models improves
  accuracy without increasing inference time,'' in \emph{International
  conference on machine learning}.\hskip 1em plus 0.5em minus 0.4em\relax PMLR,
  2022, pp. 23\,965--23\,998.

\bibitem{patel2020clustering}
E.~Patel and D.~S. Kushwaha, ``Clustering cloud workloads: K-means vs gaussian
  mixture model,'' \emph{Procedia computer science}, vol. 171, pp. 158--167,
  2020.

\bibitem{murphy2022probabilistic}
K.~P. Murphy, \emph{Probabilistic machine learning: an introduction}.\hskip 1em
  plus 0.5em minus 0.4em\relax MIT press, 2022.

\bibitem{healy2024uniform}
J.~Healy and L.~McInnes, ``Uniform manifold approximation and projection,''
  \emph{Nature Reviews Methods Primers}, vol.~4, no.~1, p.~82, 2024.

\bibitem{rashid2025novel}
J.~Rashid, S.~M. Boulaaras, M.~S. Saleem, M.~Faheem, and M.~U. Shahzad, ``A
  novel transfer learning approach for skin cancer classification on {ISIC}
  2024 {3D} total body photographs,'' \emph{International Journal of Imaging
  Systems and Technology}, vol.~35, no.~2, p. e70065, 2025.

\bibitem{hameed2024comprehensive}
M.~Hameed, A.~Zameer, and M.~A.~Z. Raja, ``A comprehensive systematic review:
  {Advancements} in skin cancer classification and segmentation using the isic
  dataset.'' \emph{CMES-Computer Modeling in Engineering \& Sciences}, vol.
  140, no.~3, 2024.

\bibitem{dolhansky2020deepfake}
B.~Dolhansky, J.~Bitton, B.~Pflaum, J.~Lu, R.~Howes, M.~Wang, and C.~C. Ferrer,
  ``The deepfake detection challenge ({DFDC}) dataset,'' \emph{arXiv preprint
  arXiv:2006.07397}, 2020.

\bibitem{huang2021deepfake}
J.~Huang, X.~Wang, B.~Du, P.~Du, and C.~Xu, ``Deepfake {MNIST+}: a deepfake
  facial animation dataset,'' in \emph{Proceedings of the IEEE/CVF
  International Conference on Computer Vision}, 2021, pp. 1973--1982.

\bibitem{ciplab2020deepfake}
CIPLab, ``Real and fake face detection dataset,'' Available:
  https://tinyurl.com/yc564acs, 2020, accessed: June 1, 2025.

\bibitem{pintelas2025quantization}
E.~Pintelas, I.~E. Livieris, and P.~Pintelas, ``Quantization-based {3D-CNNs}
  through circular gradual unfreezing for deepfake detection,'' \emph{IEEE
  Transactions on Artificial Intelligence}, 2025.

\bibitem{dobrovsky2022plant}
A.~Dobrovsky, ``Plant disease classification merged dataset,'' Available:
  https://tinyurl.com/yc89tcj3, 2022, accessed: June 1, 2025.

\bibitem{kermany2018identifying}
D.~S. Kermany, M.~Goldbaum, W.~Cai, C.~C. Valentim, H.~Liang, S.~L. Baxter,
  A.~McKeown, G.~Yang, X.~Wu, F.~Yan \emph{et~al.}, ``Identifying medical
  diagnoses and treatable diseases by image-based deep learning,'' \emph{cell},
  vol. 172, no.~5, pp. 1122--1131, 2018.

\bibitem{xu2022robust}
Y.~Xu, B.~Du, and L.~Zhang, ``Robust self-ensembling network for hyperspectral
  image classification,'' \emph{IEEE Transactions on Neural Networks and
  Learning Systems}, vol.~35, no.~3, pp. 3780--3793, 2022.

\bibitem{gu2023hierarchical}
X.~Gu, Z.~Zhang, Y.~Jiang, T.~Luo, R.~Zhang, S.~Cui, and Z.~Li, ``Hierarchical
  weight averaging for deep neural networks,'' \emph{IEEE Transactions on
  Neural Networks and Learning Systems}, vol.~35, no.~9, pp. 12\,276--12\,287,
  2023.

\bibitem{song2022learning}
H.~Song, M.~Kim, D.~Park, Y.~Shin, and J.-G. Lee, ``Learning from noisy labels
  with deep neural networks: A survey,'' \emph{IEEE transactions on neural
  networks and learning systems}, vol.~34, no.~11, pp. 8135--8153, 2022.

\bibitem{borisov2022deep}
V.~Borisov, T.~Leemann, K.~Se{\ss}ler, J.~Haug, M.~Pawelczyk, and G.~Kasneci,
  ``Deep neural networks and tabular data: A survey,'' \emph{IEEE transactions
  on neural networks and learning systems}, 2022.

\bibitem{mobilenetv3}
A.~Howard, M.~Sandler, G.~Chu, L.-C. Chen, B.~Chen, M.~Tan, W.~Wang, Y.~Zhu,
  R.~Pang, V.~Vasudevan, Q.~V. Le, and H.~Adam, ``Searching for
  {MobileNetV3},'' \emph{Proceedings of the IEEE/CVF International Conference
  on Computer Vision (ICCV)}, pp. 1314--1324, 2019.

\bibitem{zhao2025mixir}
T.~Zhao, X.~Guo, Y.~Lin, and B.~Du, ``Mixir: Mixing input and representations
  for contrastive learning,'' \emph{IEEE Transactions on Neural Networks and
  Learning Systems}, pp. 8255--8264, 2025.

\bibitem{naidu2023review}
G.~Naidu, T.~Zuva, and E.~M. Sibanda, ``A review of evaluation metrics in
  machine learning algorithms,'' in \emph{Computer Science On-line
  Conference}.\hskip 1em plus 0.5em minus 0.4em\relax Springer, 2023, pp.
  15--25.

\bibitem{garcia2010advanced}
S.~Garc{\'\i}a, A.~Fern{\'a}ndez, J.~Luengo, and F.~Herrera, ``Advanced
  nonparametric tests for multiple comparisons in the design of experiments in
  computational intelligence and data mining: Experimental analysis of power,''
  \emph{Information sciences}, vol. 180, no.~10, pp. 2044--2064, 2010.

\end{thebibliography}
% or embeded -->

\end{document}